\documentclass{article}

\usepackage{PRIMEarxiv}

\usepackage{amsmath,amssymb,amsfonts}
\usepackage{algorithmic}
\usepackage{booktabs}
\usepackage{graphicx}
\usepackage{multirow}
\usepackage[export]{adjustbox}
\usepackage[utf8]{inputenc}
\usepackage{url}
\usepackage{commath}
\usepackage{graphicx}
\usepackage{comment}
\usepackage[ruled,linesnumbered]{algorithm2e}
\usepackage{subfigure}
\usepackage{array}
\usepackage{mathtools}

\usepackage{enumitem}
\usepackage{mdframed}
\mdfsetup{linecolor=gray!70}

\usepackage{xcolor}
\usepackage{soul}

\newcommand{\best}[1]{{%
    \colorlet{foo}{pink}%
    \sethlcolor{foo}\hl{#1}}%
}

\def\tsc#1{\csdef{#1}{\textsc{\lowercase{#1}}\xspace}}
\tsc{WGM}
\tsc{QE}

\newcommand{\xy}[1]{{\color{brown}#1}}
\newcommand{\rev}[1]{{\color{black}#1}}

\DeclareMathOperator*{\dist}{dist}
\newcommand{\radius}{\text{\texttt{radius}}}

\newcommand{\minPts}{\text{\texttt{minPts}}}
\definecolor{bg}{rgb}{0.96,0.96,0.8}

\title{Fast and explainable clustering based on sorting}
\author{
  Xinye Chen \\
  Department of Mathematics \\
  The University of Manchester \\
  Manchester \\
  \texttt{xinye.chen@manchester.ac.uk} \\
   \And
  Stefan G\"{u}ttel \\
  Department of Mathematics \\
  The University of Manchester \\
  Manchester \\
  \texttt{stefan.guettel@manchester.ac.uk} \\
}

\begin{document}
\maketitle

\begin{abstract}
We introduce a fast and explainable clustering method called CLASSIX. It consists of two phases, namely a greedy aggregation phase of the sorted data into groups of nearby data points,  followed by the merging of groups into clusters. The algorithm is controlled by two scalar parameters, namely a distance parameter for the aggregation and another parameter controlling the minimal cluster size. Extensive experiments are conducted to give a comprehensive evaluation of the clustering performance on  synthetic and real-world datasets, with various cluster shapes and low to high feature dimensionality. Our experiments demonstrate that CLASSIX competes with state-of-the-art clustering algorithms. The algorithm has linear space complexity and achieves near linear time complexity on a wide range of problems. Its inherent simplicity allows for the generation of intuitive explanations of the computed clusters.
\end{abstract}


\maketitle

\section{Introduction}
Clustering is a widely-used unsupervised learning technique to find {patterns and structure} in data. Clustering algorithms group the data points into distinct clusters such that points within a cluster share similar characteristics on the basis of their distance, density or other spatial properties, while points in two distinct clusters are less similar. 
Applications of clustering methods are wide-ranging,  including areas like finance~\cite{RePEc:fip:fedawp:2004-20}, traffic~\cite{10.1145/1162678.1162679}, civil engineering~\cite{DEOLIVEIRA2011380}, and bioinformatics~\cite{10.1093/bioinformatics/btaa613}. 
Clustering algorithms can be categorized in different ways, but for our purposes distance-based and density-based  clustering methods are the most relevant. Distance-based clustering methods, such as k-means~\cite{1056489}, consider the pairwise distance between points in order to decide whether they should be contained in the same cluster. Density-based clustering algorithms, such as DBSCAN \cite{10.5555/3001460.3001507}, take a more global view by assuming that data occurs in continuous areas of high density surrounded by low-density regions. A key advantage of many density-based clustering algorithms is that they can handle arbitrarily shaped clusters without specifying the number of clusters in advance. On the other hand, they typically require more parameter tuning.


We propose a novel clustering method called CLASSIX\footnote{CLASSIX is a contrived acronym of ``CLustering by Aggregation with Sorting-based Indexing'' and the letter ``X'' for ``explainability.''} which shares features with both distance and density-based methods.  The method comprises two phases: \emph{aggregation} and \emph{merging}. In the aggregation phase, data points are sorted along their first principal component  
\cite{Hotelling1933AnalysisOA} 
and then grouped using a greedy aggregation technique. The sorting is key to traversing the data with almost linear complexity, provided that the number of pairwise distance computations is small. While the initial sorting requires an average-case complexity of $O(n\log n)$, 
it is only performed on scalar values irrespective of the dimensionality of the data points. As a consequence, the cost of this initial sorting is almost negligible compared to computations on the full-dimensional data. As we will demonstrate experimentally, an almost linear complexity is indeed observed in practice. The aggregation phase is followed by the merging of overlapping groups into clusters using either a distance or density-based criterion. The density-based merging criterion usually results in slightly better clusters than the distance-based criterion, but the latter has a significant speed advantage. CLASSIX depends on two parameters and its tuning is relatively straightforward. In brief, there is a distance parameter $\radius$ that serves as a tolerance for the grouping in the aggregation phase, while an $\minPts$ parameter specifies the smallest acceptable cluster size. This is similar to the parameters used in DBSCAN but, owing to the initial sorting of the data points, CLASSIX does not perform spatial range queries for each data point. 

Our comparisons demonstrate that  CLASSIX achieves excellent performance against popular methods like \texttt{\texttt{k-means++}}, meanshift, DBSCAN, $\text{HDBSCAN}^{\ast}$  and Quickshift++ with respect to speed and several metrics of cluster quality. In addition, CLASSIX's inherent simplicity allows for the easy explainability of the clustering results. The Python code available at 
\begin{center}
    \url{https://github.com/nla-group/classix}
\end{center}
implements CLASSIX with all these features, including the ability to query textual explanations for the clustering. \rev{The full data and precise choice of hyperparameters for each method and each single experiment can be found in the same repository (subfolder \texttt{exp}).}

The rest of this paper is structured as follows.  Section~\ref{Section: Related work} reviews some of the existing popular clustering algorithms which share features with CLASSIX. Section~\ref{sec:classix} introduces the CLASSIX algorithm, starting with the greedy aggregration approach and a discussion of two different merging techniques. We also show how to treat outliers and out-of-sample data. Section~\ref{sec:theory} is devoted to theoretical considerations of the computational complexity of CLASSIX and how its performance depends on parameters such as the data dimension, Section~\ref{Section: EX} contains comprehensive performance tests of CLASSIX  and comparisons with other popular clustering algorithms. Concluding remarks and a discussion of possible future work are contained in section~\ref{Section: Conclusion}. Finally, the Appendix gives an overview of using the Python implementation and a demonstration of the explainability feature.

\section{Related work}\label{Section: Related work}
CLASSIX shares features with at least two classes of clustering methods, namely distance-based and density-based methods. Here we briefly review some of the most important algorithms in these classes, highlighting the common features shared with CLASSIX and also their differences. 

Two of the most popular distance-based clustering algorithms are k-means \cite{1056489} and k-medoids \cite[Chp. 2]{doi.org/10.1002/9780470316801.ch2}. 
These algorithms  partition the data into $k$ clusters by iteratively relocating the cluster centers and reassigning data points to their nearest center. As a consequence, these distance-based methods tend to form clusters of spherical shapes and it is generally hard to cluster data with more complicated cluster shapes~\cite{Jain_2010}. In terms of computational complexity, it is rather hard to predict how slow or fast k-means performs~\cite{10.1145/1137856.1137880}, but variants like \texttt{k-means++}~\cite{10.5555/1283383.1283494}, mini-batch k-means \cite{10.1145/1772690.1772862}, or the contribution in \cite{10.5555/3041838.3041857} can significantly improve on the original algorithm in practice. Both k-means and k-medoids require the user to specify~$k$, the expected number of clusters, and this is often not easy in practice. \rev{Another distance-based clustering method is Principal Direction Divisive Partitioning (PDDP) by \cite{Boley1998}, which proceeds by repeatedly splitting clusters along the  first principal direction of the current cluster. PDDP invokes multiple principle component computations whereas CLASSIX performs only one. Furthermore, as the splitting of data is done by separating hyperplanes, PDDP is best suited for clusters that can be contained in disjoint  convex polytopes.  There is also an improved variant called dePDDP~\cite{TASOULIS20103391} which is capable of estimating the number of clusters automatically.}

Density-based clustering methods do not \rev{typically require a priori} knowledge of the number of clusters. Among these methods are the popular mean shift \cite{400568, 1000236} and DBSCAN \cite{10.5555/3001460.3001507} algorithms. These methods rely on the property that the density of points at the ``core'' of a cluster is higher than in its surrounding. Mean shift  performs clustering by iteratively shifting data points to nearby peaks (modes) of the empirical density function, using kernel density estimation. Data associated with the same mode are assigned to a cluster.  Quick shift~\cite{Vedaldi2008QuickSA} provides a faster alternative of mean shift clustering using Euclidean medoids as shifts. The recently introduced Quickshift++  further improves  clustering performance by first estimating the data modes and providing  guarantees that data are assigned to their appropriate cluster mode \cite{pmlr-v80-jiang18b}. 

The DBSCAN method  essentially relies on two parameters, namely the neighborhood radius and a minimum number of data points in every neighborhood. The original implementation uses an R* tree data structure to perform range queries, and the performance of DBSCAN relies crucially on the efficient implementation of such a tree structure. 
Other trees can be used as well, such as \rev{$k$-d tree} \rev{or} ball tree, but as can be seen in \cite{kriegel2017black} the implementation of such structures is nontrivial. 
As pointed out by \cite{10.1145/2723372.2737792}, DBSCAN has a worst-case time complexity of $O(n^2)$ and a time complexity of $\Omega(n^{4/3})$ if the number of features is greater than~three. This problem often is referred to as the curse of dimensionality~\cite{10.1145/276698.276876, v008a014}.




Many variants of DBSCAN have been proposed to mitigate the  aforementioned issues. OPTICS computes an augmented cluster ordering over a range of parameters and thus provides a versatile basis for both automatic and interactive cluster analysis \cite{10.1145/304182.304187, Hahsler2019-wa}. \cite{Gunawan} proposes a  DBSCAN variant for two-dimensional data which has a theoretical  time complexity of $O(n \log n)$. To fix the limitation left by \cite{Gunawan}, \cite{10.1145/2723372.2737792} introduce the concept of  $\rho$-approximate DBSCAN with an expected linear time complexity $O(n)$ for a small sacrifice in accuracy. 
However, \cite{10.1145/3068335} claim that the original DBSCAN can be competitive with $\rho$-approximate DBSCAN for reasonable parameters choices and effective index structures. There are also  parallel implementations like PDSDBSCAN \cite{Patwary2012ANS}.  APSCAN \cite{CHEN2011973} is a parameter-free clustering method that uses affinity propagation  \cite{Frey07AffinityPropagation} to infer the local density of a dataset. In such a way, APSCAN can cluster datasets with varying density and preserve the associated nonlinear data structure.  HDBSCAN \cite{10.1007/978-3-642-37456-2_14} and $\text{HDBSCAN}^{\ast}$ \cite{10.1145/2733381}, for simplicity just referred to as ``HDBSCAN'' here, extend DBSCAN to hierarchical clustering algorithms and then extract a flat clustering  based on the stability of clusters. Again, this allows for  variable-density clusters and, possibly, a more  robust parameter selection. DBSCAN++ \cite{pmlr-v97-jang19a} is a modification of DBSCAN based on the observation that density estimates for merely a subset of $m \ll n$  data points need to be computed, resulting in a speedup of DBSCAN while achieving similar clustering results as DBSCAN. To select the $m$ data points, uniform and greedy $k$-center-based sampling approaches are proposed. 

CLASSIX shares features all of the aforementioned methods. CLASSIX  begins with the greedy aggregation of the data into groups, each of which is identified with a so-called starting point. Similar to the neighborhood centers in DBSCAN, the starting points can be interpreted as a reduced-density estimator of the data. However, in CLASSIX the groups are also re-used to cluster the data, not just as estimators. CLASSIX's two-stage approach also shares similarities with hierarchical clustering methods like HDBSCAN. In the merging phase of CLASSIX, a distance-based criterion can be used, e.g., based on the Euclidean distance. However, as opposed to the distance-based methods discussed above, the merging phase of CLASSIX is non-iterative and does not require a specification of the number of clusters. Instead, the number of clusters is determined by the algorithm based on a $\texttt{radius}$ parameter  and the minimum number of data points a cluster,  $\minPts$. While we use a similar notation for these parameters as DBSCAN, their meaning in CLASSIX is slightly different. The groups formed in CLASSIX are not necessarily spherical as the scanned neighborhoods in DBSCAN. Instead, they are formed from a starting point by traversing the data in the sorted order and only including points that have not been aggregated before. The $\minPts$ criterion is applied on the cluster level in CLASSIX, while in DBSCAN it is used to distinguish between noise, core,  and boundary points.

\section{The CLASSIX algorithm}\label{sec:classix}

In this section we introduce the CLASSIX algorithm, starting with a discussion of the initial preparation of the data  (section~\ref{sec:sort}),  followed by a discussion of the aggregation (section~\ref{sec:aggr}) and merging (section~\ref{sec:merge}) phases. The short sections~\ref{sec:outl} and \ref{sec:outofs} discuss the treatment of outliers and out-of-sample data, respectively.

\subsection{Data preparation}\label{sec:sort}

Let us assume that we are given $n$ raw data points $p_1, \ldots, p_n \in \mathbb{R}^{d}$ which we would like to cluster. Throughout this work all vectors are column vectors and we assume that $d\ll n$. The dimensionality~$d$ is also referred to as the number of features. In the following, we will denote the processed data points as $x_1,\ldots, x_n$. 

As a first step, we will center all data points by taking off the empirical mean value of each feature:
\[
    x_i := p_i - \mathrm{mean}(\{p_j\}).
\]
This operation has a complexity of $O(dn)$, i.e.\ linear in $n$, and it can be performed \emph{in-place}, meaning that the data points $p_i$ can be overwritten with the mean-centered values $x_i$ if memory consumption is an issue. 

The second step consists of computing the first principal component $v_1\in \mathbb{R}^{d}$, i.e., the vector along which the data $\{ x_i\}$ exhibits largest empirical variance. This vector can be computed by  a thin singular value decomposition of the tall-skinny data matrix $X := [x_1,\ldots, x_n]^T \in\mathbb{R}^{n\times d}$,
\begin{equation}\label{eq:svd}
    X = U \Sigma V^T, 
\end{equation}
where $U\in \mathbb{R}^{n\times d}$ and $V\in \mathbb{R}^{d\times d}$ have orthonormal columns and $\Sigma =\mathrm{diag}(\sigma_1,\ldots, \sigma_d)\in \mathbb{R}^{d\times d}$ is a diagonal matrix such that 
$
\sigma_1 \geq \sigma_2 \geq \cdots \geq \sigma_d \geq 0.
$
The principal components are given as the columns of $V=[v_1,\ldots,v_d]$ and we require only the first column $v_1$. 
The score of a point $x_i$ along $v_1$ is
\[
    \alpha_i := x_i^T v_1 = (e_i^T X) v_1 = (e_i^T U \Sigma V^T)v_1 = e_i^T u_1 \sigma_1,
\]
where $e_i$ denotes the $i$-th canonical unit vector in $\mathbb{R}^n$. In other words, the scores $\alpha_i$ of all points can be read off from the first column of $U = [u_1,\ldots, u_d]$ times $\sigma_1$.
The computation of the scores using a thin SVD  requires $O(n d^2)$ operations and is therefore linear in~$n$ \cite[chapter~8.6]{gene2013}. 

The third (and crucial) step is to order all data points $x_i$ by  their $\alpha_i$ scores; that is,
\[
    ( x_i ) := \mathrm{sort}(\{ x_i \})
\]
so that $\alpha_1 \leq \alpha_2 \leq \cdots \leq \alpha_n$ with each $\alpha_i = x_i^T v_1$. Note that $v_1$ is not uniquely determined even when $\sigma_1 > \sigma_2$ as also $-v_1$ is a principal component, but it does not matter if the order of the $\alpha_i$ is reversed from largest to smallest. The sorting generally requires $O(n\log n)$ operations on average, and at most $O(n)$ memory, provided that an efficient algorithm such as Quicksort~\cite{hoare1962quicksort}, IntroSort~\cite{introsort} or TimSort~\cite{timsort} is used. It is important to highlight again that the sorting is performed on \emph{scalar} values $\alpha_i$, irrespective of the data dimensionality~$d$. This will result in a significant advantage in performance over tree-based query structures such as R*
or ball tree, which work with the $d$-dimensional data points and are nontrivial to implement efficiently; see, e.g., the discussion in~\cite{kriegel2017black}.

Finally, we estimate the \emph{median extend} of the data as the smallest number $\texttt{mext}>0$ such that there are $\lceil n/2 \rceil$ data points with a score $\alpha_i$ in the interval $[-\texttt{mext}, \texttt{mext}]$. The use of this parameter is merely to make any distance computations independent of the data scale,  ensuring that approximately half of the data points $x_i$ have a score $\alpha_i$ that lie within a ball of radius $\texttt{mext}$ about the origin. As the scalars $\alpha_i$ are already sorted, $\texttt{mext}$ can be computed in $O(n)$ operations. An alternative and better estimate is  $\texttt{mext}:= \mathrm{median}(\{\| x_i \|\})$, but this is slightly more  expensive to compute as it involves $n$~Euclidean norm computations for a cost of $O(nd)$ operations, and another sorting of scalar norm values costing $O(n\log n)$. 

We note that throughout this work we will use the Euclidean norm $\|\,\cdot\,\|$ for any distance computations. If a weighted Euclidean needs to be used (giving different weight to different features), this weighting can simply be applied to the original data points $\{p_i\}$ before performing the follow-up steps as described above.

\subsection{Aggregation phase}\label{sec:aggr}

The aim of CLASSIX's aggregation phase is to group nearby data points together in a computationally inexpensive manner, where ``nearby'' is simply defined in terms of the Euclidean  distance $\mathrm{dist}(x_i,x_j) = \| x_i  - x_j \|$. We consider two data points $x_i$ and $x_j$ close if  $$\mathrm{dist}(x_i,x_j)\leq \radius \cdot \texttt{mext} =: R.$$ Here, $\radius$ is a user-chosen tolerance and \texttt{mext} is the median extend estimated from the data, as explained in the previous section.

The greedy aggregation now proceeds as follows: starting with the first data point $x_1$ (i.e., the one with the smallest $\alpha_i$~value), we assign the data points $x_j$ for $j=2,3,\ldots$ to the same group $G_1$ if $\mathrm{dist}(x_1,x_j)\leq R$. If done in a naive way, this would   require computing all $n-1$ distances between the $x_j$ and $x_1$. However, we can utilize the Cauchy--Schwarz inequality 
\begin{equation}\label{eq:lower}
 | \alpha_i - \alpha_j | = | v_1^T x_i - v_1^T x_j | \leq   \| x_i - x_j \| = \dist(x_i, x_j) 
\end{equation}
to avoid doing this: as we have sorted the $x_i$ such that $\alpha_1\leq \alpha_2 \leq \cdots \leq \alpha_n$, we know that
\[
\text{if} \  \alpha_j - \alpha_i > R\  \text{for some $j>i$} \ \text{then}  \   \dist(x_i, x_k) > R \ \text{for all $k\geq j$}.
\]
Hence, we can terminate adding data points to the first group $G_1$ as soon as we encounter a data point $x_j$ such that $\alpha_j - \alpha_1 > R$. The point $x_1$ is also referred to as the \emph{starting point} of group $G_1$.

Once the first group $G_1$ is completed, the assignment process continues with the first data point $x_i$ that is not part of $G_1$. This point $x_i$ will be the starting point of the next group $G_2$. (We emphasize that the starting points do not need to be searched for; they are always the first data point that was skipped when forming the previous group.) As before, we add unassigned data points $x_j$, $j>i$, to the  group $G_2$ until we encounter a point such that $\alpha_j - \alpha_i > R$. And this process continues until all points are assigned to a group.

A pseudocode of the greedy aggregation algorithm is given in Algorithm~\ref{algo:aggr}. Once completed, all $n$ data points have been partitioned into groups $G_1,\ldots,G_\ell$ with their corresponding starting points $x_{s(1)}, \ldots, x_{s(\ell)}$.

\LinesNumberedHidden{
\begin{algorithm}[ht]
\caption{Greedy aggregation of data points into groups}\label{algo:aggr}
\hspace*{-5mm}\begin{minipage}[h]{\textwidth}
\begin{enumerate}[itemsep=-2pt]
    \item Given  data points $x_1,x_2,\ldots,x_n$ sorted such that $\alpha_1\leq \alpha_2\leq \cdots\leq \alpha_n$.\\ 
	Label all of them as ``unassigned''.
	\item Let $x_i$ be the first unassigned point $x_i$ (referred to as a ``starting point'').\\ 
	If there are no unassigned points left, terminate.
	\item FOR $j=i+1 : n$
	\item \quad Compute $d_{ij}:=\dist(x_i,x_j)$
	\item \quad IF $d_{ij}\leq R$ and $x_j$ is unassigned,  
		 assign $x_j$ to the same group as $x_i$ 
    \item \quad ELSE IF $\alpha_j - \alpha_i > R$, go to Step 2.
\end{enumerate}
\end{minipage}
\end{algorithm}}

\subsection{Merging phase}\label{sec:merge}

\begin{figure}[ht]
\centering
\includegraphics[width=0.62\textwidth]{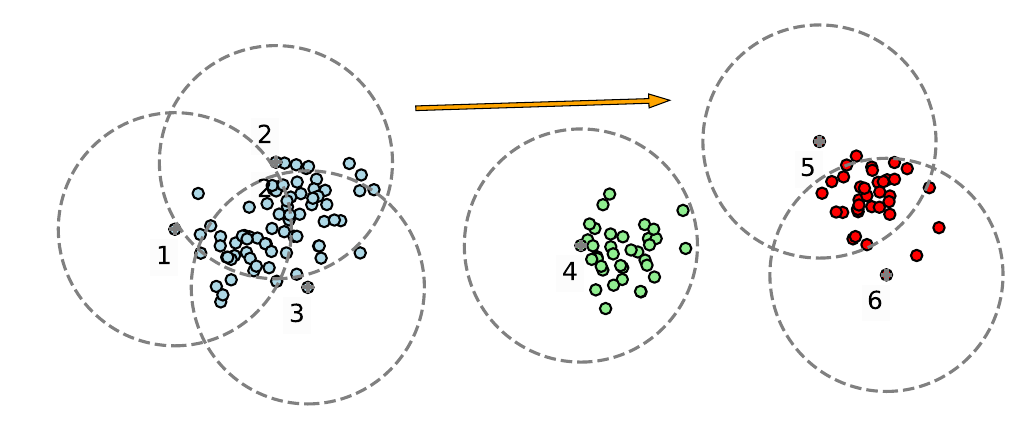}
\caption{Illustration of a point cloud aggregated into $\ell=6$ groups (dashed circles) and merged into $k=3$ clusters (labelled by colour). The data points are ordered along the direction of the first principal component indicated by the arrow, and the six starting points of the groups (which are also data points) follow that same order.}
\label{fig:scheme}
\end{figure}

The aim of this phase is to merge the groups $G_1,\ldots,G_\ell$ into clusters $C_1,\ldots, C_k$, $k\leq \ell$. See also Figure~\ref{fig:scheme} for illustration. We discuss two approaches for doing this, one based solely on the distance of the starting points $x_{s(1)}, \ldots, x_{s(\ell)}$, and the other based on the density of the points in the groups and a  volume of intersection.

\subsubsection{Distance-based merging}

We first highlight that the points $x_i$ assigned to each group $G_j$ are still sorted by their first principal coordinates $\alpha_i$ and, as a consequence, so are the starting points; i.e., $\alpha_{s(1)} \leq \cdots \leq \alpha_{s(\ell)}$. 

In the distance-based merging approach we will merge two groups $G_i$ and $G_j$ if the distance of their starting points satisfies
\begin{equation}\label{eq:critmerg}
    \mathrm{dist}(x_{s(i)}, x_{s(j)}) \leq \texttt{scale} \cdot R,
\end{equation}
where \texttt{scale} is a tuning parameter in the interval $[1,2]$, which can be set to $\texttt{scale}=1.5$ in most cases. 
Note that two distinct starting points $x_{s(i)}$ and $x_{s(j)}$, $i<j$, will never have a distance that is less than $R$, as otherwise they would have ended up in the same group $G_i$ during aggregation. Also, if two starting points have a distance greater than $2R$, their surrounding $R$-balls have no overlap. Let us the denote the $R$-ball surrounding a point $x$ as 
\[
    \mathcal{B}(x, R) := \{ y\in\mathbb{R}^d : \mathrm{dist}(x, y) \leq R \}.
\]
The above requirement that $\texttt{scale}\in [1,2]$ ensures that there is a nonempty intersection of the balls $\mathcal{B}(x_{s(i)}, R)$  and $\mathcal{B}(x_{s(j)}, R)$.

To determine which pairs of starting points satisfy the merging criterion \eqref{eq:critmerg}, we can again use the termination trick based on the corresponding principal coordinates. Given a starting point $x_{s(i)}$, we inspect the starting points that succeed it, namely $x_{s(i+1)}, x_{s(i+2)}, \ldots$ in that order. If any of these starting points is a distance not more than $\texttt{scale}\cdot R$ away from $x_{s(i)}$, it should be in the same cluster as $x_{s(i)}$. As soon as we encounter a starting point $x_{s(i+k)}$ with the property $\alpha_{s(i+k)} - \alpha_{s(i)} >\texttt{scale}\cdot R$ we can terminate the search. 

Once all starting points $x_{s(i)}$ have been traversed, we effectively have a list of pairs of starting points that satisfy the merging criterion~\eqref{eq:critmerg}. These pairs can be thought of as the edges of a directed graph. We then simply determine clusters of starting points by computing the connected components of this graph (or its undirected version) using, e.g., depth-first search~\cite{hopcroft1973algorithm} or  a disjoint-set data structure  \cite{10.1145/364099.364331}. The complexity of this procedure is $O(\max(\ell, E))$, where $E$ is the number of edges in the graph. All data points $x_i$ which are not starting points are naturally assigned to the same cluster as the starting point of the group they belong to.

\subsubsection{Density-based merging}\label{sec:dense}
In density-based merging we will join two starting points $x_{s(i)}$ and $x_{s(j)}$ (and the points in their associated groups $G_i$ and $G_j$, respectively) into one cluster if the density of the points that lie in the intersection $\mathcal{B}(x_{s(i)},R) \cap \mathcal{B}(x_{s(j)},R)$ is at least as big as the overall density of the two groups. Using $|\,\cdot\,|$ to denote cardinality of a set and  $\mathrm{vol}(\,\cdot\,)$ to denote the volume of a set, this can be formalized as follows: $x_{s(i)}$ and $x_{s(j)}$ belong to the same cluster if 
\begin{equation}\label{eq:critdense}
    \frac{|\mathcal{B}(x_{s(i)},R) \cup  \mathcal{B}(x_{s(j)},R)|}{\mathrm{vol}  \left(\mathcal{B}(x_{s(i)},R) \cup  \mathcal{B}(x_{s(j)},R)\right) } 
    \leq 
    \frac{|\mathcal{B}(x_{s(i)},R) \cap  \mathcal{B}(x_{s(j)},R)  |}{\mathrm{vol}  \left(\mathcal{B}(x_{s(i)},R) \cap  \mathcal{B}(x_{s(j)},R)\right) }.
\end{equation} 

The volume of the intersection of two $R$-balls in $\mathbb{R}^d$ is 
\[
\mathrm{vol}  \left(\mathcal{B}(x,R) \cap  \mathcal{B}(y,R)\right) 
= \frac{\displaystyle \pi^{d/2} \cdot R^{d} \cdot I_{1-\mathrm{dist}(x,y)^2/(4R^2)}(d/2 + 1, 1/2)}{\Gamma(d/2 + 1)},
\]
with the  incomplete beta function defined as 
\[
    I_s(a,b) = \frac{\Gamma(a + b)}{\Gamma(a)\Gamma(b)}\int_{0}^{s}t^{a-1}(1 - t)^{b-1}\,\mathrm{d}t, \quad s \in [0,1]; 
\]
see \cite{Li2011ConciseFF}. The volume of an $R$-ball is given as $\mathrm{vol} \left(\mathcal{B}(x,R)\right) = \mathrm{vol}  \left(\mathcal{B}(x,R) \cap  \mathcal{B}(x,R)\right)$, from which the volume of the union of two $R$-balls is easily obtained.
Two $R$-balls can only have a nonempty overlap if their centers are not more than $2R$ apart. Hence we can again limit the number of pairs $G_i$ and $G_j$, $i<j$, for which to check the merging criterion \eqref{eq:critdense} by using the principal coordinate-based criterion: if $\alpha_{s(j)} -  \alpha_{s(i)}> 2R$, the groups $G_i$ and $G_j$ do not need to be inspected for intersection.

As in the distance-based case, the set of all pairs of starting points that satisfy \eqref{eq:critdense} can be thought of as the edges of a graph, and we  determine clusters of starting points by computing the connected components of this graph. All data points $x_i$ which are not starting points are naturally assigned to the same cluster as the starting point of the group they belong to.

\smallskip

\noindent \textbf{Remark:} As is well known, the volume of an $R$-ball tends to zero as the dimension~$d$ increases, 
\[
    \mathrm{vol}(\mathcal{B}(x,R)) \sim \frac{1}{\sqrt{d\pi}} \left(\frac{2\pi \mathrm{e}}{d}\right)^{d/2} R^d.
\]
Similarly, the volume of any intersection of two balls tends to zero, and even the relative volume of the intersection of two balls (i.e., the ratio of the intersection volume and the volume of a ball) tends to zero unless both balls have the same centers. As a consequence, density-based clustering is only suitable for low feature dimensions.

\subsection{Treatment of outliers}\label{sec:outl}

After all groups have been merged into $k$ clusters $C_1,\ldots,C_k$, we optionally apply a removal process for outliers. In CLASSIX, outliers are simply defined as the points in clusters that have a small number of elements, namely clusters such that $|C_j| < \texttt{minPts}$. There are two options for processing outliers.
\begin{itemize}
    \item \textbf{Reassign outliers to larger clusters:} All points in a cluster $C_j$ with less than \texttt{minPts} points will be assigned to larger nearby clusters. This is done on the group level using the starting points. Every group $G_i$ that was merged into $C_j$ has an associated starting point $x_{s(i)}$. We simply find a starting point $x_{s(i')}$ nearest to $x_{s(i)}$ that belongs to a cluster $C_{j'}$ with at least \texttt{minPts} points, and reassign $G_i$ to cluster $C_{j'}$. This procedure will succeed to eliminate all outliers if there is at least one cluster  with \texttt{minPts} or more points. Otherwise, all clusters will remain outliers.
    \item \textbf{Return outliers separately:} All points in clusters  with less than \texttt{minPts} points will be labelled as outliers and returned as such to the user.
\end{itemize}

\begin{figure}[ht]
	\centering
	\includegraphics[width=0.36\textwidth]{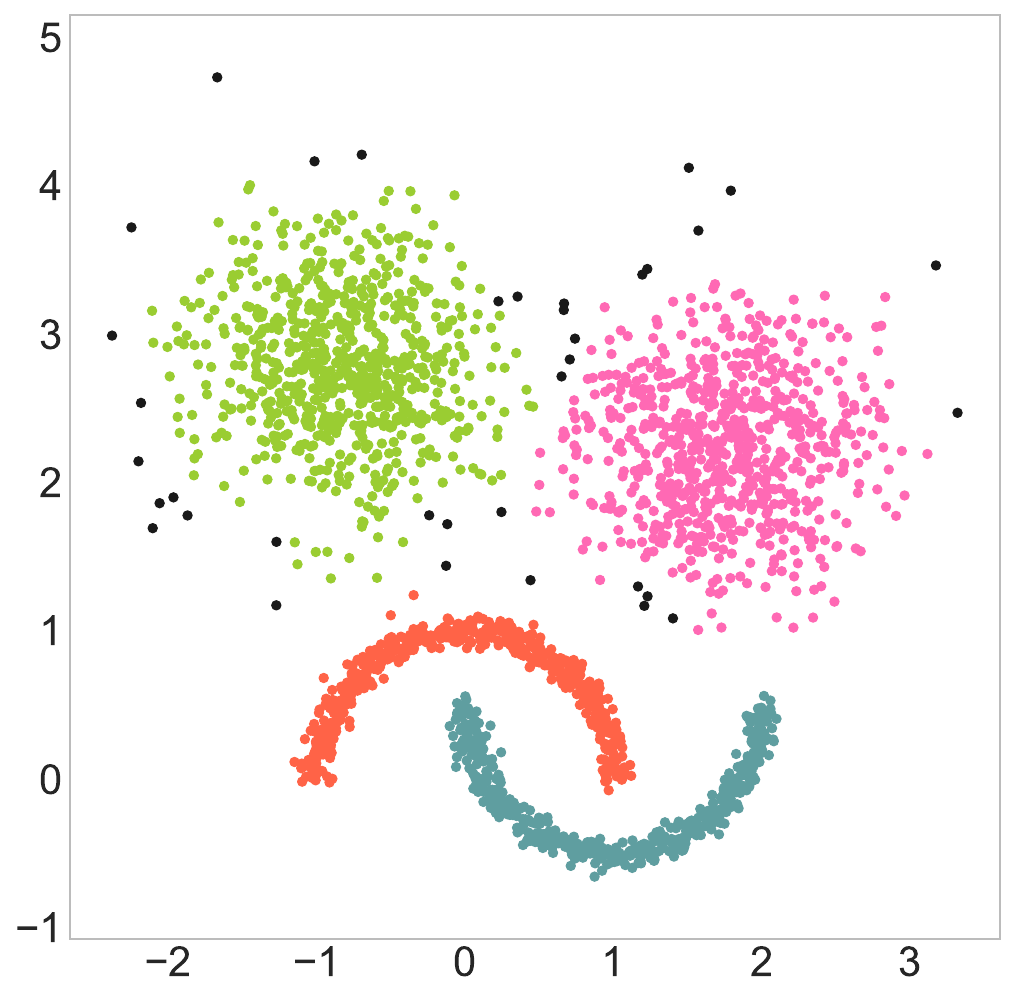}
	\includegraphics[width=0.36\textwidth]{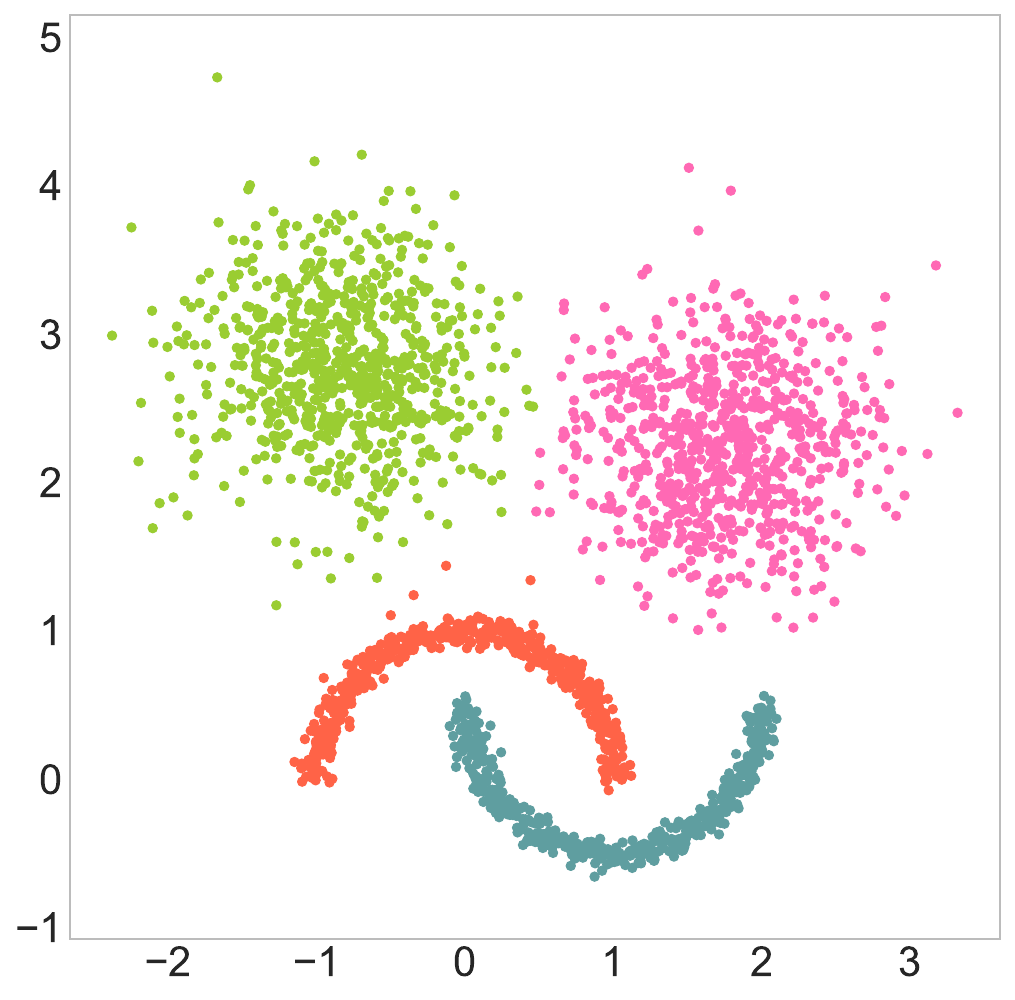}
	\caption[Treatment of outliers]{Demonstration of the outlier treatment. We apply CLASSIX to synthetic data with $n=2500$ points using the parameters $R=0.1$ and $\minPts=5$. Left: Without outlier reassignment, 27 clusters are identified, four of which are genuine ground-truth clusters (shown in colour) and each of the other 23~clusters have less than $\minPts$ points (shown as the black dots). 
	Right: The reassignment of the 23 outlier clusters results in four clusters of good quality.}
	\label{fig:noises}
\end{figure}

We illustrate these two options applied to an example with synthetic data in Figure~\ref{fig:noises}. 

\subsection{Out-of-sample data}\label{sec:outofs}

\begin{figure}[ht]
	\centering
	\subfigure{\includegraphics[width=0.3\textwidth]{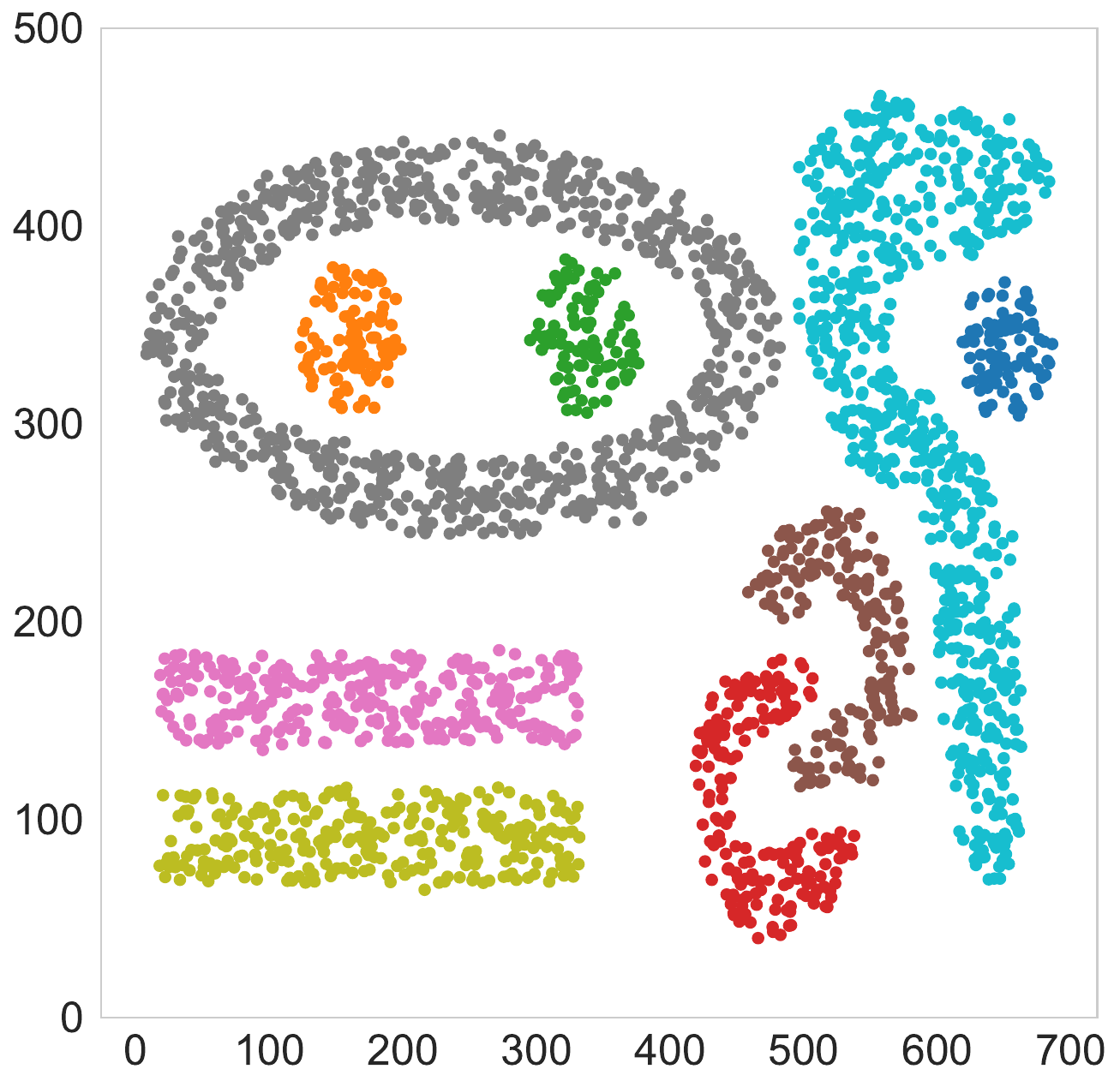}}\qquad
	\subfigure{\includegraphics[width=0.3\textwidth]{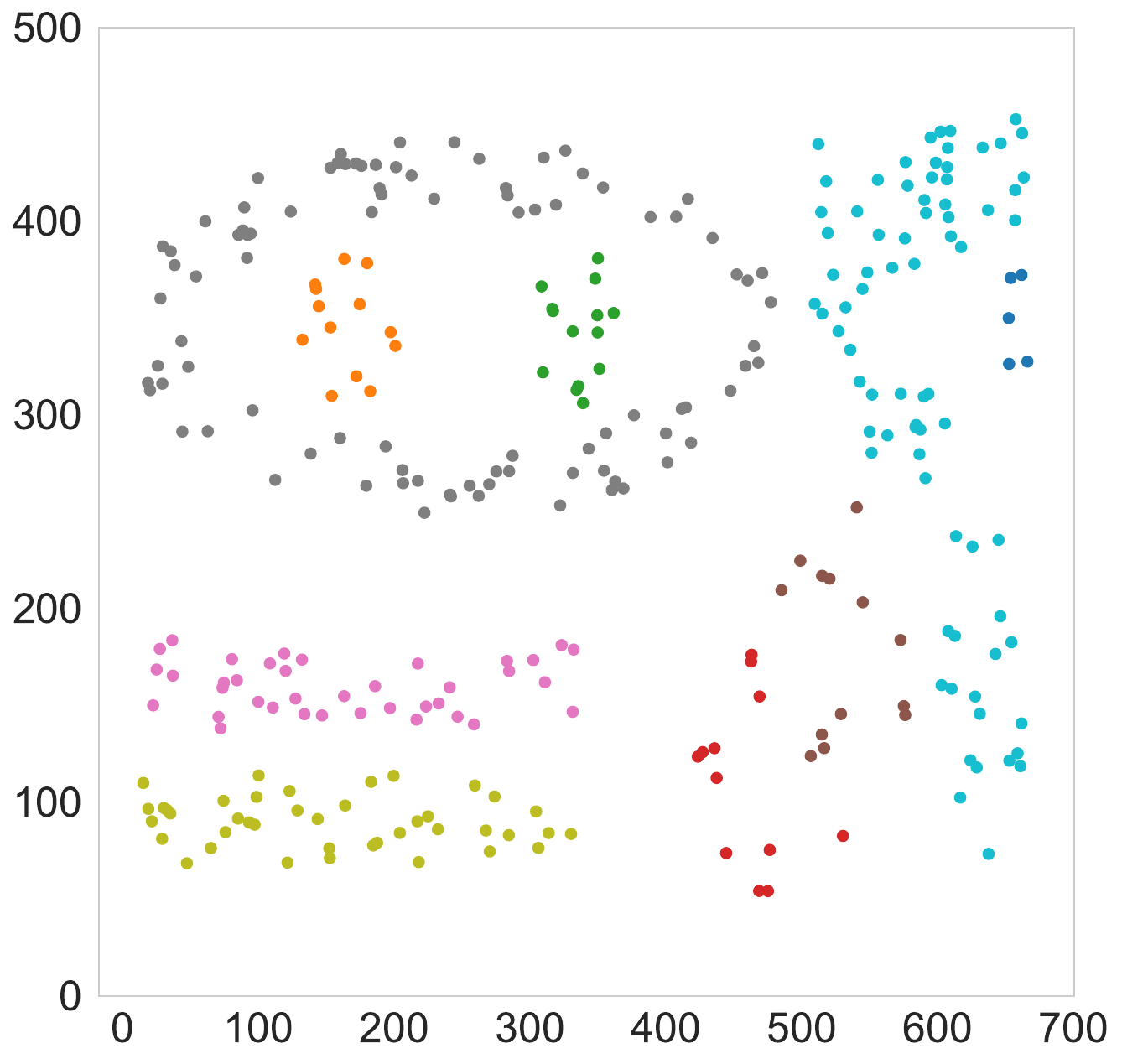}}\\
	\subfigure{\includegraphics[width=0.3\textwidth]{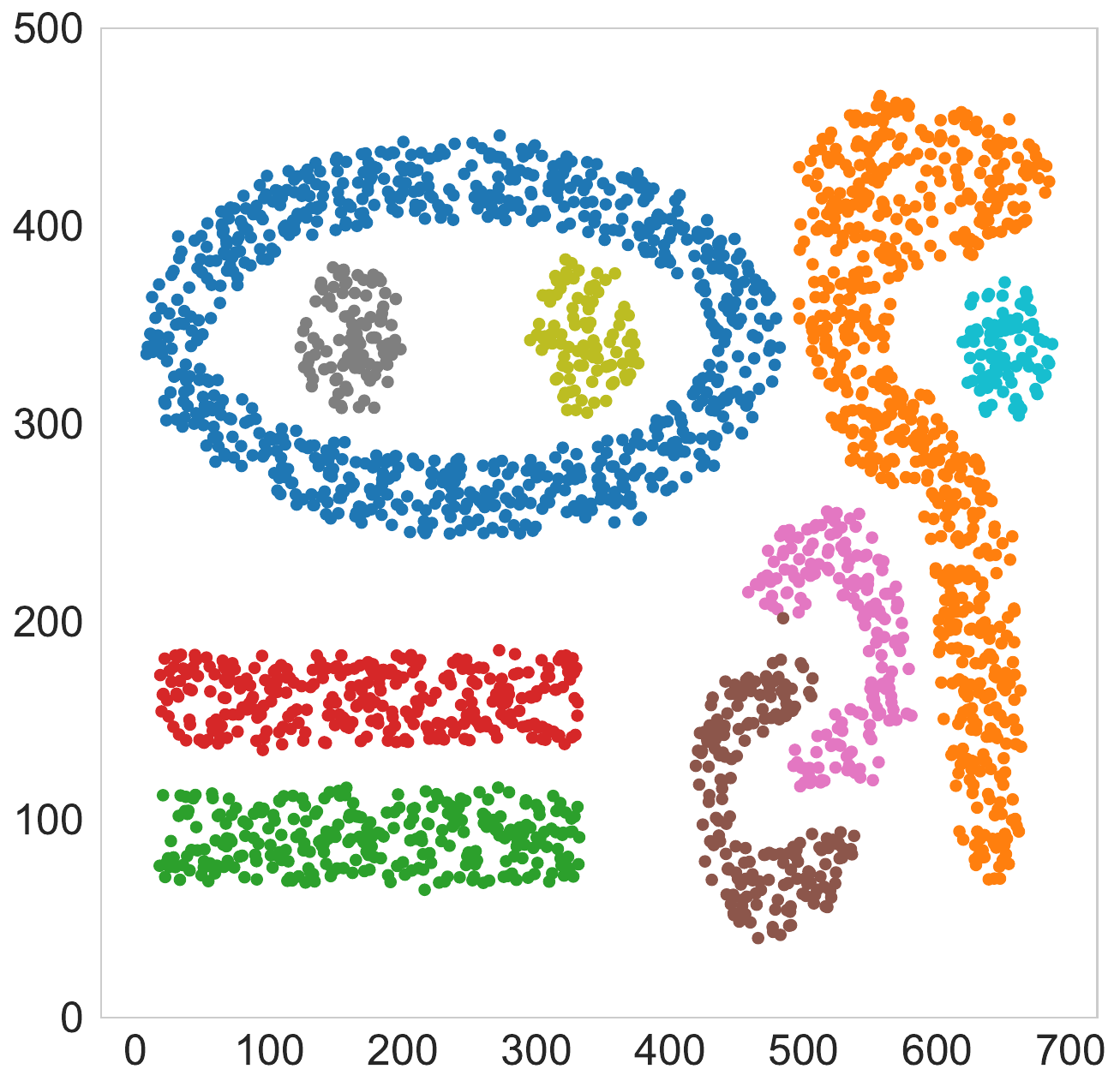}}\qquad
	\subfigure{\includegraphics[width=0.3\textwidth]{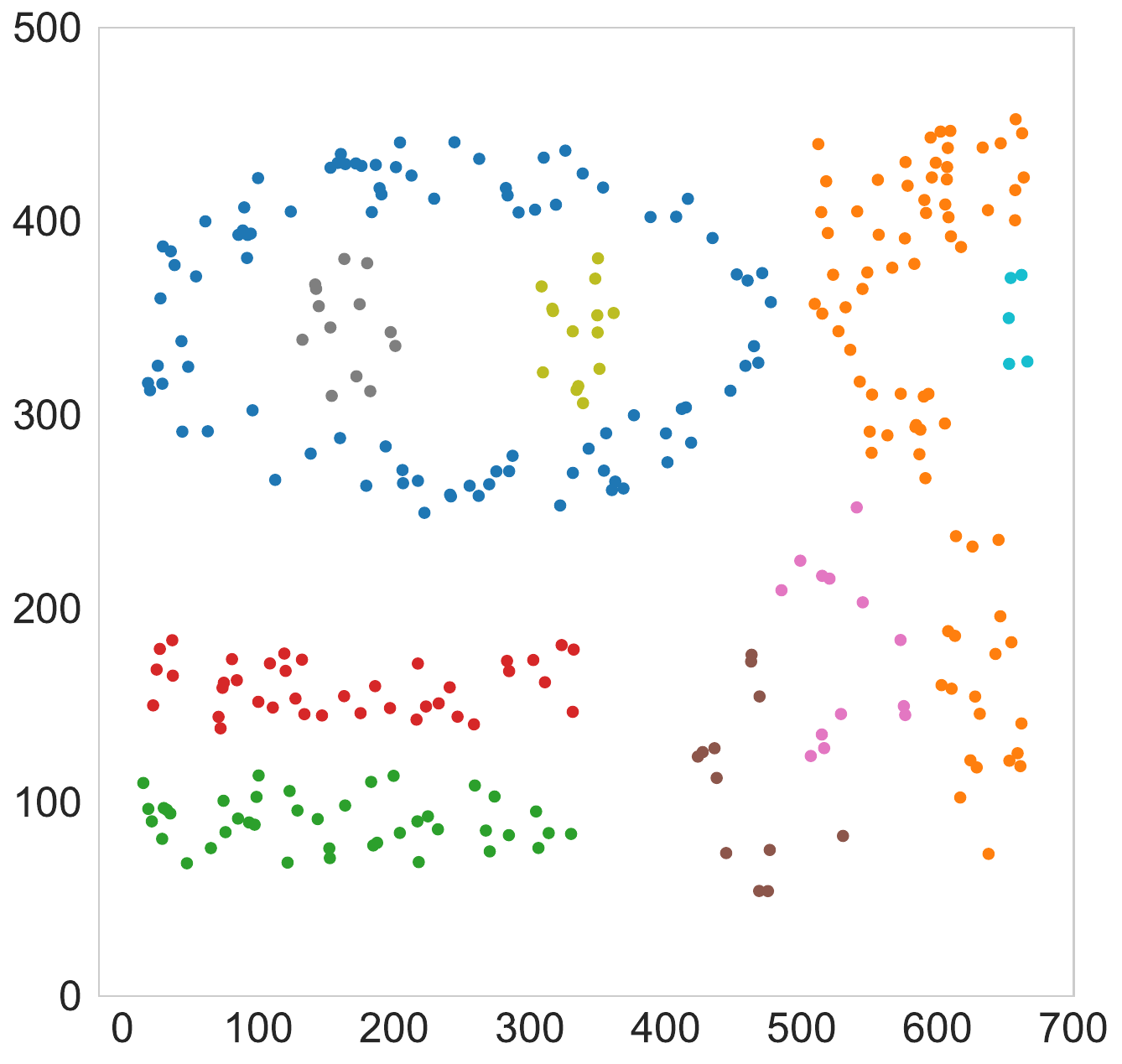}}
	\caption{Illustration of CLASSIX clustering with out-of-sample data. Top: Ground truth labels of the train and test data (left and right, respectively). Bottom: CLASSIX clustering of the train and test data (left and right, respectively). The train vs test split is 90\% vs 10{\%}.} 
	\label{fig:ncomplex9}
\end{figure}

The reassignment strategy for outliers described in section~\ref{sec:outl} can also be used to classify out-of-sample data based on a previously  trained model. That is, CLASSIX will allocate a new data point to its closest group (measured by distance to starting points) and then assign it to the associated cluster. This is illustrated in \figurename~\ref{fig:ncomplex9}. In this example, CLASSIX is first run on a train set (90\% of the overall data points) to obtain the clusters, and then tested on the remaining 10\% of the data (left vs right plots). In this example, both the train and test data are clustered with 100\% accuracy when compared to the provided ground truth (top vs bottom plots).

\section{Theoretical considerations}\label{sec:theory}

\subsection{Efficiency}
The efficiency of the aggregation phase  crucially depends on the number of pairwise distance calculations that are performed in Step~4 of~Algorithm~\ref{algo:aggr}. In the best case, each point $x_i$ is involved in exactly one such computation, resulting in $O(n)$ distance computations overall. In the worst case, the distances between each point $x_i$ and all points $x_{i+1},\ldots,x_n$ that succeed it are computed. This amounts to 
\[
    (n-1) + (n-2) + \cdots + 2 = \frac{(n-1)n}{2} - 1 = O(n^2)
\]
distance computations, an unacceptably high number in most practical situations. The hope is that through the sorting of the data points $x_i$ by their principal coordinates $\alpha_i$,  the early termination criterion in Step~6 of Algorithm~\ref{algo:aggr} is triggered often enough so that, on average, the number of distance computations a data point is involved in remains very small. 

An unassigned data point $x_j$ is involved in more than one distance computation only if there is a starting point $x_i$, $i<j$, such that  $\alpha_j - \alpha_i \leq R$ but $\mathrm{dist}(x_i,x_j)>R$. So, it is natural to ask the following question: 

\smallskip

\quad \quad \emph{How likely is it that $|\alpha_i - \alpha_j|\leq R$, yet $\mathrm{dist}(x_i,x_j)>R$?}

\smallskip

We first note that, using the singular value decomposition~\eqref{eq:svd} of the data matrix $X$, we can derive an upper bound on $\mathrm{dist}(x_i,x_j)$ that complements the lower bound \eqref{eq:lower}. Using that $x_i^T = e_i^T X = e_i^T U \Sigma V^T$ and denoting the elements of $U$ by $u_{i,j}$, we have 
\begin{equation*}
\begin{aligned}
\|x_i - x_j\|^2 &=  |\alpha_i - \alpha_j|^2 + \left\|\, \big[(u_{i2}-u_{j2}) ,\ldots, (u_{id}-u_{jd})\big] \begin{bmatrix} \sigma_2 & & \\
 & \ddots & \\  & & \sigma_d\end{bmatrix}\,\right\|^2 \\ 
&\le |\alpha_i - \alpha_j|^2 + \|u_i - u_j\|^2 \cdot \left\|\,\begin{bmatrix} \sigma_2  & &\\
 & \ddots & \\ & & \sigma_d\end{bmatrix}\,\right\|^2  \\
&\le |\alpha_i - \alpha_j|^2 + 2 \sigma_2^2.
\end{aligned}
\end{equation*}
Therefore, 
\begin{equation}\label{eq:sigma2}
    |\alpha_i - \alpha_j |^2 \leq \mathrm{dist}(x_i, x_j)^2 \leq |\alpha_i - \alpha_j|^2 + 2 \sigma_2^2,
\end{equation}
and the gap in these inequalities is determined by $\sigma_2$, the second singular value of~$X$. Indeed, if $\sigma_2=0$, then all data points $x_i$ lie on a straight line passing through the origin and their distances correspond exactly to the difference in their principal coordinates. This is a best-case scenario for Algorithm~\ref{algo:aggr}, as points $x_j$  will just be added in their sorted order to a group until the distance from the starting point $x_i$ exceeds $R$, in which case that $x_j$ becomes the next starting point and a new group is formed. As no point needs to be revisited, the number of pairwise distance computations is smallest possible. 
If, on the other hand, $\sigma_2$ is relatively large compared to $\sigma_1$, the gap in the inequalities becomes  large as well and the $\alpha_i,\alpha_j$ may not be a good proxy for the point distance $\dist(x_i,x_j)$.

\subsection{Parameter dependency}\label{sec:simpmod}
It is clear that there is no natural order of data points in two or higher dimensions, and it is expected that some data points will be involved in more than one distance computation. In order to get a qualitative understanding of how the number of distance computations depends on the various parameters (dimension $d$, singular values of the data matrix, group radius $R$, etc.),  we consider the following  model. Let $\{x_i\}_{i=1}^n$ be a large sample of points whose $d$ components are normally distributed with zero mean and standard deviation $[1,s,\ldots,s]$, $s<1$, respectively. These points describe an elongated ``Gaussian blob'' in $\mathbb{R}^d$, with the elongation controlled by~$s$. In the large data limit $(n\to\infty)$ the singular values of the data matrix $X=[x_1,\ldots,x_n]^T$ approach $\sqrt{n},s\sqrt{n},\ldots,s\sqrt{n}$ and the principal components approach the canonical unit vectors $e_1,e_2,\ldots,e_d$. As a consequence, the principal coordinates $\alpha_i = e_1^T x_i$ follow a standard normal distribution, and hence for any $c\in\mathbb{R}$ the probability that $|\alpha_i - c|\leq R$ is given as 
\[
 P_1 =  \displaystyle  P_1(c,R) = \frac{1}{\sqrt{2\pi}} \int_{c-R}^{c+R}  e^{-r^2/2}\, \mathrm{d}r.
\]
On the other hand, the probability that $\|x_i - [c,0,\ldots,0]^T\|\leq R$ is given by
\[
P_2 = P_2(c,R,s,d) =  \frac{1}{\sqrt{2\pi}}  \int_{c-R}^{c+R} e^{-r^2/2}  \cdot {F\left(\frac{R^2 - (r-c)^2}{s^2}; d-1\right)}\, \mathrm{d}r,
\]
where $F$ denotes the $\chi^2$ cumulative distribution function. In this model we can think of the point $[c,0,\ldots,0]^T$ as a starting point, and our aim is to identify all data points $x_i$ within a radius $R$ of this starting point.

Since $\|x_i - [c,0,\ldots,0]^T\|\leq R$ implies that $|\alpha_i - c|\leq R$, we have $P_1 \geq P_2$. Hence, the quotient $P_2/P_1$ can be interpreted as a conditional probability of a point $x_i$ satisfying $\|x_i - [c,0,\ldots,0]^T\|\leq R$ given that $|e_1^T x_i - c|\leq R$, i.e.
\[
P = P\big(\|x_i - [c,0,\ldots,0]^T\|\leq R\, \big| \, |e_1^T x_i - c|\leq R\big) = P_2/P_1.
\]
Ideally, we would like this quotient $P$ be close to $1$. As all the points $x_i$ are sorted by their first principal coordinates~$\alpha_i$, all points within a radius of $R$ from $[c,0,\ldots,0]^T$ (belonging to the same group) are very likely to have consecutive indices if $P\approx 1$. On the other hand, if $P$ is very small, then the sorting has very little effect of indexing nearby points close to each other.

It is now easy to study the dependence of the quotient $P=P_2/P_1$ on the various parameters.
First note that $P_1$ does not depend on $s$ nor $d$, and hence the only effect these two parameters have on $P$ is via the factor $F\left(\frac{R^2 - (r-c)^2}{s^2}; d-1\right)$ in the integrand of $P_2$. This term corresponds to the  probability that the sum of squares of~$d-1$ independent Gaussian random variables with mean zero and standard deviation~$s$ is less or equal to $R^2 - (r-c)^2$. Hence, $P_2$ and therefore~$P$ are monotonically decreasing as $s$ or $d$ are increasing. This is consistent with intuition: as $s$ increases, the elongated point cloud $\{x_i\}$ becomes more spherical and hence it gets more difficult to find a direction in which to enumerate the points naturally. And this problem gets more pronounced in higher dimensions~$d$.

We now show that $P$ converges to $1$ as $R$ increases. First note that for an arbitrarily small $\epsilon>0$ there exists a radius $R_1>1$ such that $P_1(c,R_1-1) > 1-\epsilon$. Further, there is a $R_2>1$ such that 
\[
F\left(\frac{R_2^2 - (r-c)^2}{s^2}; d-1\right) > 1-\epsilon \quad \text{for all \quad $r\in[c-R_2+1, c+R_2-1]$}.
\]
To see this, note that the cumulative distribution function $F$ monotonically increases from $0$ to $1$ as its first argument increases from $0$ to $\infty$. Hence there exists a value $T$ for which $F(t,d-1)>1-\epsilon$ for all $t\geq T$. Now we just need to find $R_2$ such that 
\[
\frac{R_2^2 - (r-c)^2}{s^2} \geq T \quad \text{for all}\quad  r \in [c-R_2+1, c+R_2-1].
\]
The left-hand side is a quadratic function with roots at $r=c-R_2$ and $r=c+R_2$ and symmetric with respect to the maximum at $r=c$. Hence choosing $R_2$ such that
\[
\frac{R_2^2 - ([c+R_2-1]-c)^2}{s^2} = T, \quad \text{i.e.,} \quad R_2 = \left(\frac{Ts^2 + 1}{2}\right)^{1/2},
\]
or any value $R_2$ larger than that, will be sufficient. Now, setting $R=\max \{R_1,R_2\}$, we have
\[
P_2 \geq \frac{1}{\sqrt{2\pi}}  \int_{c-R+1}^{c+R-1} e^{-r^2/2}  \cdot {F\left(\frac{R^2 - (r-c)^2}{s^2}; d-1\right)}\, \mathrm{d}r \geq (1-\epsilon)^2.
\]
Hence, both $P_1$ and $P_2$ come arbitrarily close to $1$ as $R$ increases, and so does their quotient $P_2/P_1$.

\begin{figure}[h]
\centering
\includegraphics[width=16cm]{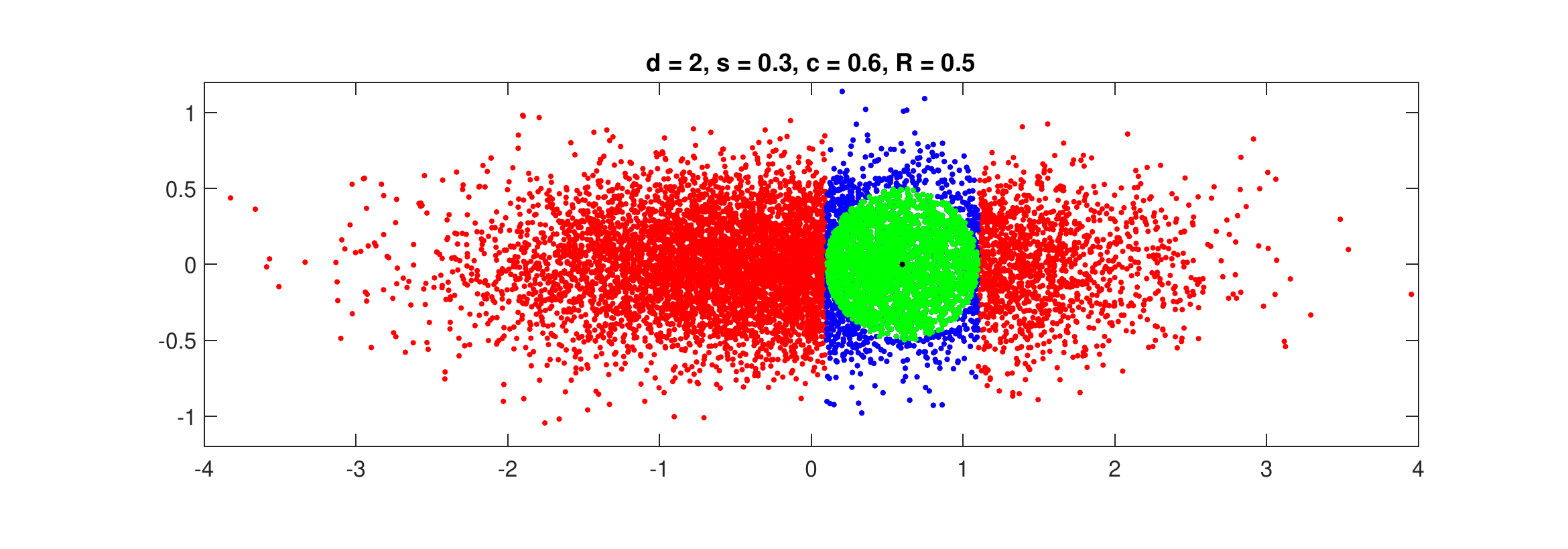}\\
\includegraphics[width=8.5cm]{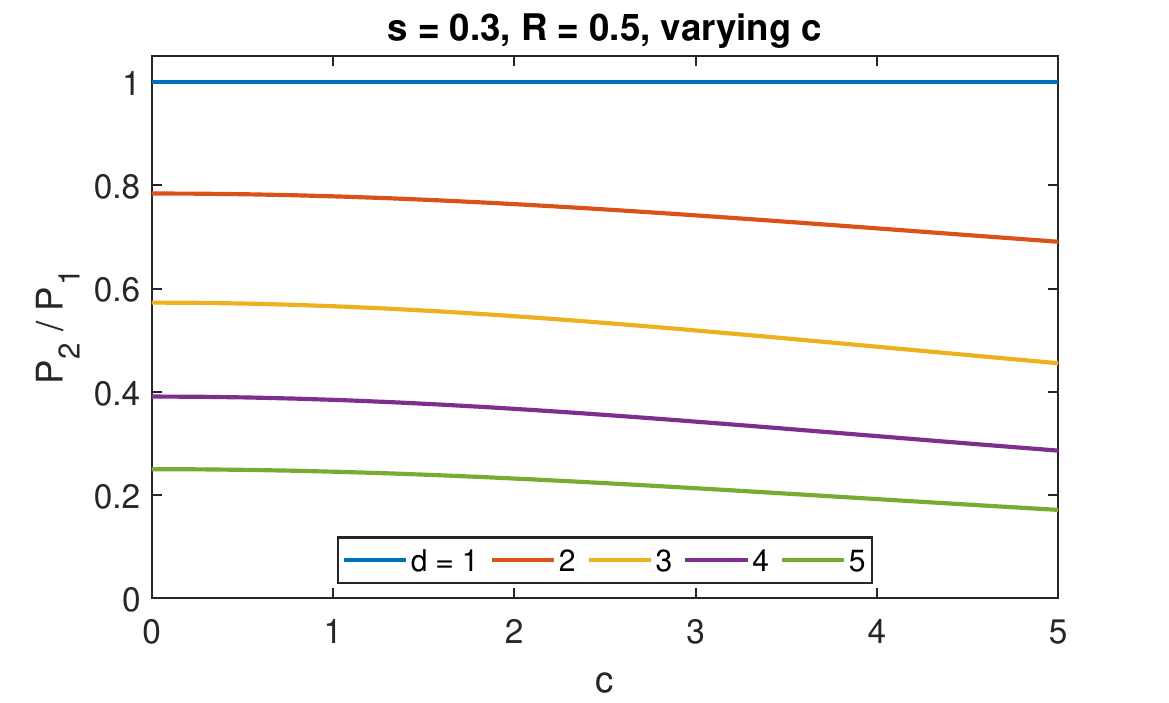}\hspace*{-5mm}
\includegraphics[width=8.5cm]{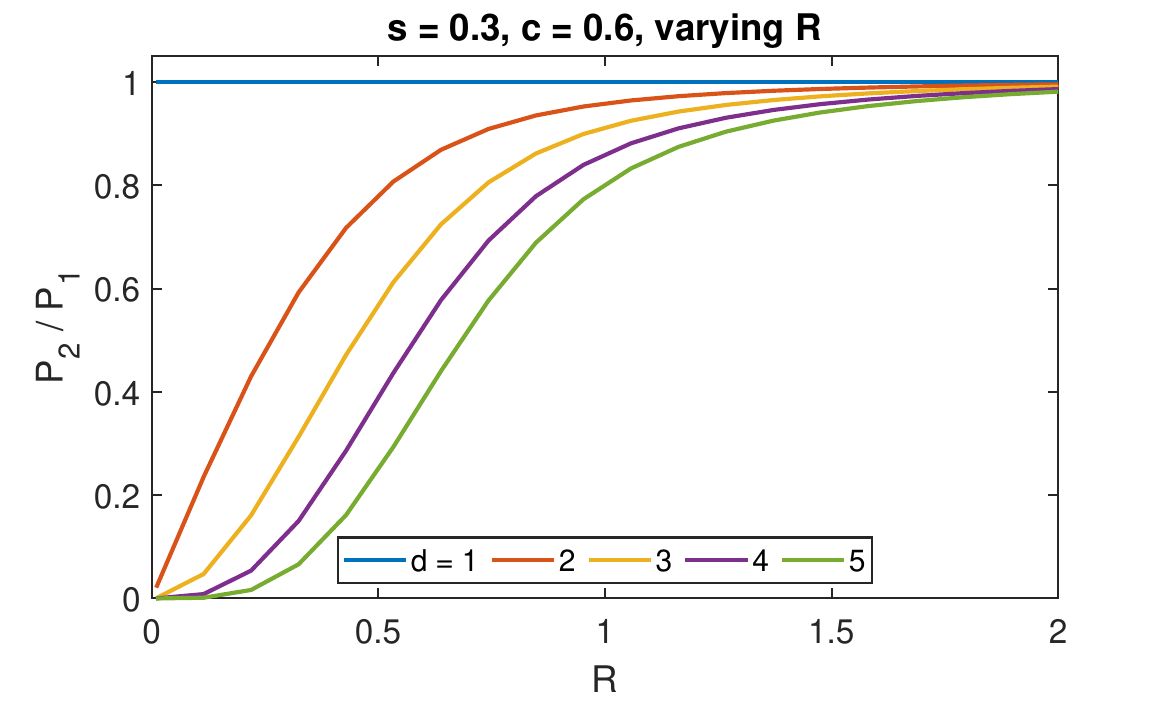}
\caption{Illustration of the  model analyzed in section~\ref{sec:simpmod}. Top: A Gaussian blob of $n=10^4$ points (red dots) in $d=2$ dimensions, with the horizontal components having a standard deviation of $1$ and the vertical components having a standard deviation of $s=0.3$. The radius for the aggregation is chosen as $R=0.5$. The blue dots are all points whose first component (first principal coordinate) is within a distance of~$R$ from $c=0.6$, while the \rev{green} points are within a distance of $R$ from the group center $c=[0.6,0]^T$ (shown as the black dot). Bottom left: The quotient $P_2/P_1$ corresponding to the ratio of green and blue points as a function of $c$ and $d$. A ratio closer to $1$ means that sorting-based aggregation is more efficient, with fewer points being revisited. Bottom right: $P_2/P_1$ as a function of $R$ and $d$.}
\label{fig:gausstest}
\end{figure}

We illustrate our findings in Figure~\ref{fig:gausstest}. The  model just analyzed only gives \rev{qualitative} insights as it does not take into account that our aggregation algorithm (i) uses starting points that are not necessarily at the center of a group and (ii) removes points once they have been assigned to a group. Nevertheless, we can read off from  Figure~\ref{fig:gausstest} that aggregation generally gets harder in higher dimensions and when $R$ gets smaller.

\section{Experiments}\label{Section: EX}
In this section we conduct extensive experiments on various datasets to evaluate the performance of CLASSIX and compare it to other clustering algorithms. The quality of a clustering will be evaluated using metrics like the adjusted Rand index (ARI) and the adjusted mutual information (AMI); see  \cite{Hubert1985ComparingP} and  \cite{vinh2010information}, respectively. \rev{Additionally, we report the Fowlkes-Mallows index (FMI)~\cite{doi:10.1080/01621459.1983.10478008} and the V-Measure (VM)~\cite{rosenberg-hirschberg-2007-v} for our tests with the UCI Machine Learning Repository, scikit-learn benchmark and the shape clusters test.} For a fairest possible comparison of timings we select widely-used clustering algorithms for which reliable and tuned implementations are publicly available. In particular, we use Cython compiled implementations of \texttt{k-means++}, DBSCAN, HDBSCAN, and Quickshift++. For the HDBSCAN and Quickshift++ implementations we refer to \cite{mcinnes2017accelerated} and \cite{pmlr-v80-jiang18b}, respectively, while the other  methods are implemented in the scikit-learn library~\cite{scikit-learn}. All reported runtimes are averaged timings of ten consecutive runs. For the reproducibility of our results, all datasets are chosen from publicly available sources or generated with open-source software.  All computations are done on a Dell PowerEdge R740 Server with two Intel Xeon Silver 4114 2.2G processors, 1,536 GB RAM, and 1.8~TB disk space.  All algorithms are forced to run in one thread.

\subsection{Gaussian blobs}

In this first test we study the sensitivity of several clustering algorithms to the data size~$n$ and the feature dimensionality~$d$. To this end we generate $n$~data points in $d$~dimensions forming ten isotropic Gaussian blobs of unit standard deviation using the \texttt{sklearn.datasets.make\_blobs} function in  scikit-learn.

\subsubsection{Dependence on data size}
We compare \texttt{k-means++}, DBSCAN (using the index query structures ``balltree'' and ``$k$-d tree''), HDBSCAN, and CLASSIX. The data are $n$~points forming ten isotropic Gaussian blobs  of unit standard deviation with $n$ varying between $5,000$ and $50,000$ while the feature dimension $d=10$ is fixed.  The method's hyperparameters are chosen as follows. For CLASSIX, we set $\radius=0.3$ and $\minPts=5$. For DBSCAN, we use $\epsilon = 3$ and $\minPts = 1$. Quickshift++ is run with $k=20$ and $\beta=0.7$. Finally, \texttt{k-means++} is run with the ground-truth value of $k=10$ clusters. The results are presented in the top two plots of \figurename~\ref{fig:size_dim_test}, with the ARI shown on the left and the runtime shown on the right. All the curves are plotted with a 95\% confidence band estimated from the ten consecutive runs of each method. 

\begin{figure}[ht]
\centering
\subfigure{\includegraphics[width=0.42\textwidth]{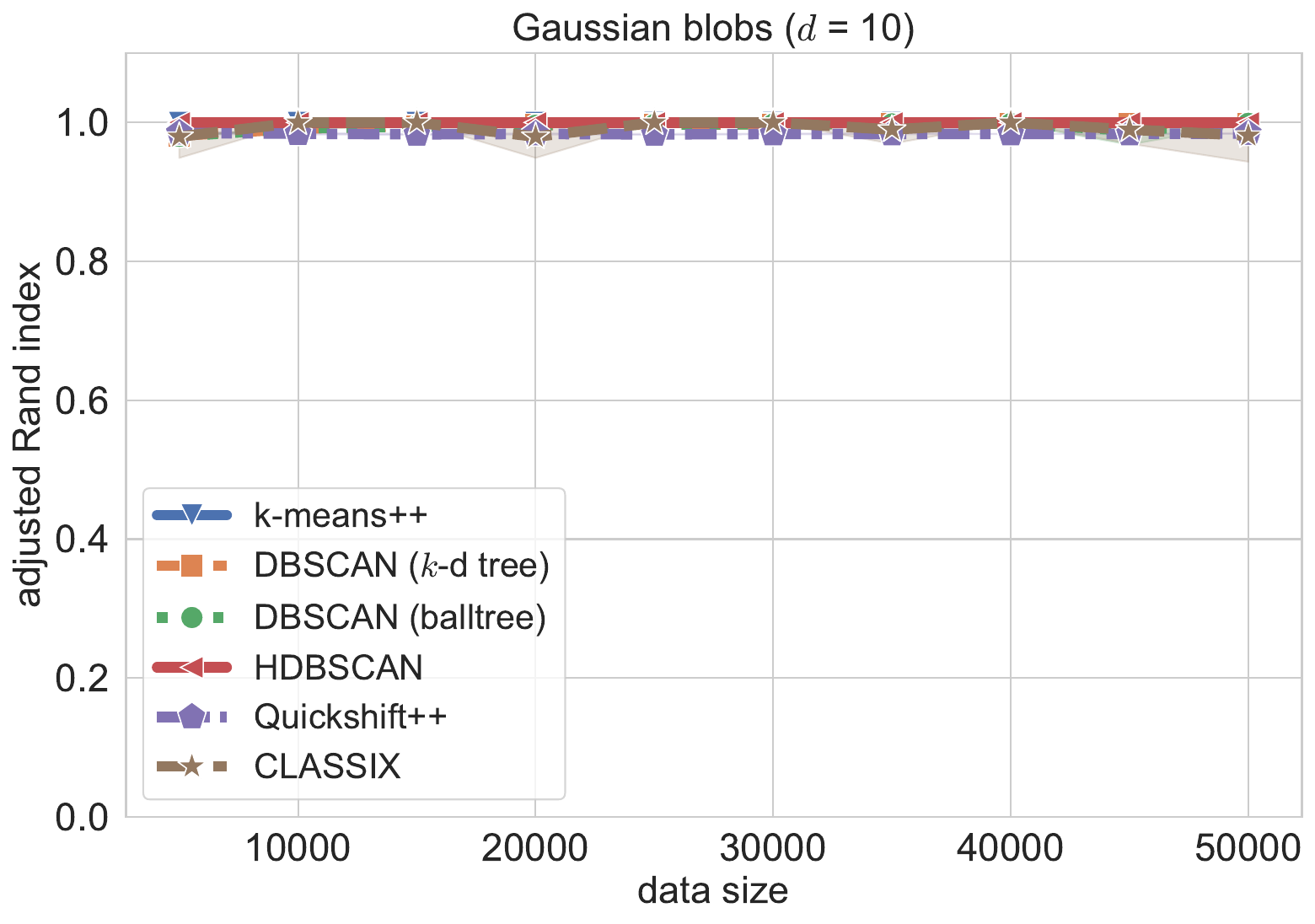}}
\ \ \subfigure{\includegraphics[width=0.418\textwidth]{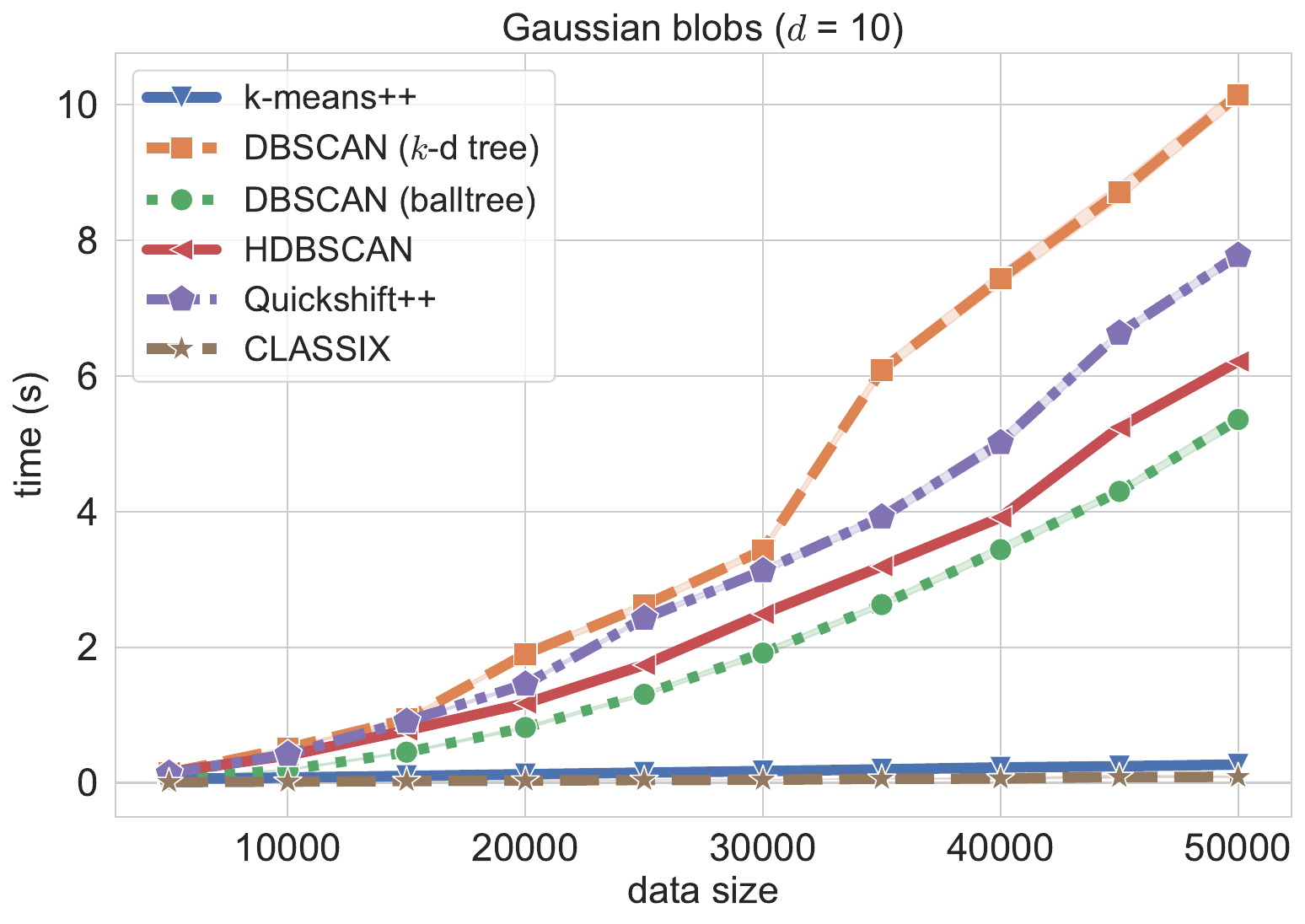}}\\
\centering\subfigure{\includegraphics[width=0.42\textwidth]{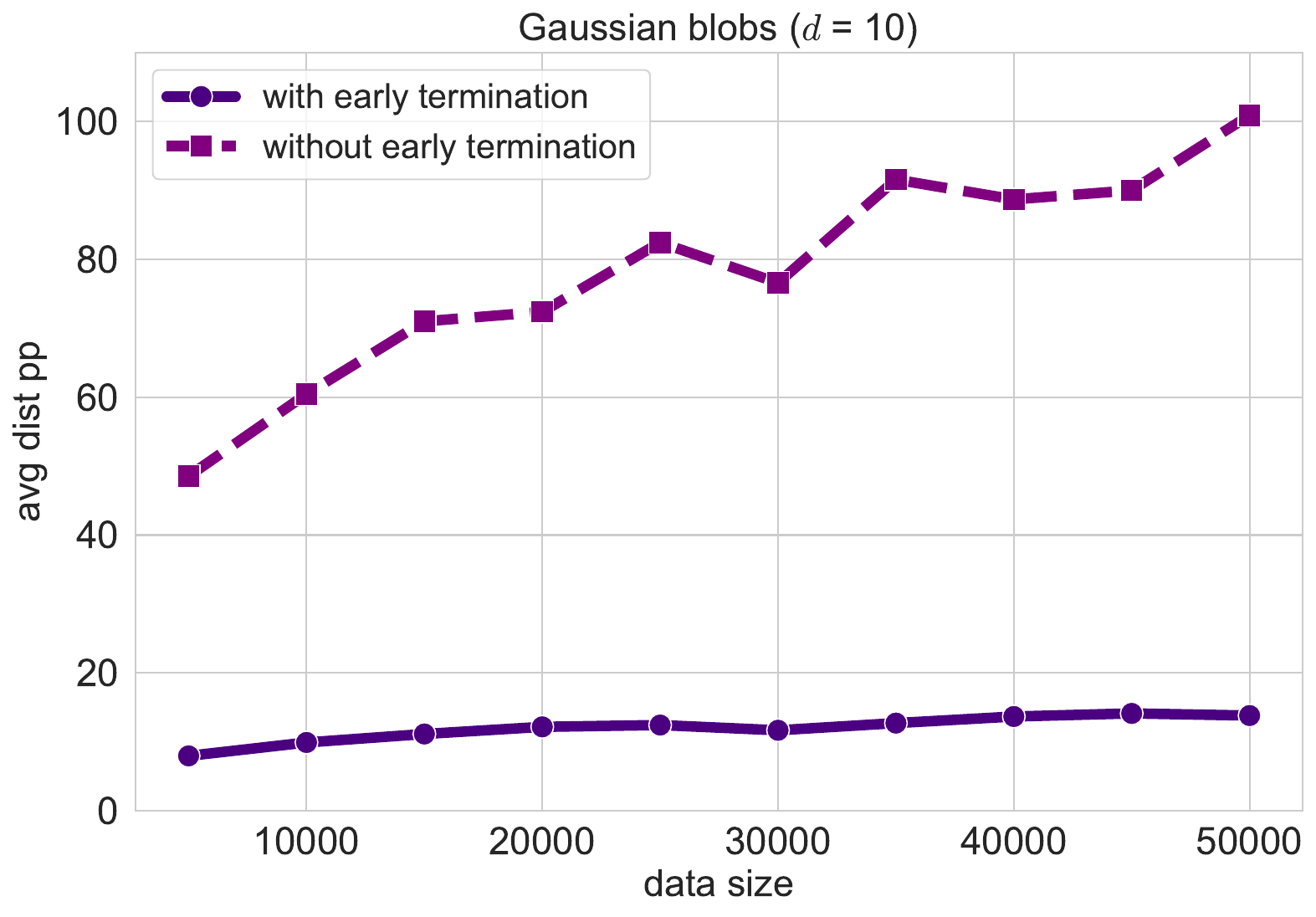}}
\caption[Scalability test]{Performance evaluation of various clustering methods on the Gaussian blobs data. Top: We vary the number of data points  from $n=5,000$ to $50,000$ and measure the adjusted Rand index (left) and timing (right) for each method. The feature dimension is fixed at $d=10$ in all cases. Bottom: Average number of distance calculations per data point in the CLASSIX aggregation phase with and without early termination.}
\label{fig:size_n_test}
\end{figure}

We find that all compared methods obtain almost perfect clustering results in this test as judged from an ARI score close to one; see  \figurename~\ref{fig:size_n_test} (top left). In terms of timings, both \texttt{k-means++} and CLASSIX achieve very good performance, with the overall computation time increasing approximately linearly as $n$ increases; see \figurename~\ref{fig:size_n_test} (top right). The other methods exhibit a nonlinear, possibly quadratic, increase in runtime. \rev{(We found that these observations remain qualitatively the same when the number of Gaussian blobs is varied, e.g.~to 5 or 20.)}

The slow, seemingly linear, increase in runtime of CLASSIX is owed to the small number of distance computations between data points due to the early termination condition discussed in section~\ref{sec:aggr}. To confirm this, we show in \figurename~\ref{fig:size_n_test} (bottom) the \emph{average number of distance calculations per data point} (\texttt{avg\_dist\_pp}) as $n$~increases, both with and without the early termination condition. We find that early termination leads to a significant reduction in the number of distance calculations. In fact, for this data \texttt{avg\_dist\_pp} appears to stay bounded with early termination as~$n$ increases. 

\subsubsection{Dependence on data dimension}
We now use a fixed number of $n=10,000$ samples of 10 Gaussian blobs, and vary the feature dimension from $d=10$ to 100. The methods' hyperparameters are the same as in the previous section, except that for DBSCAN we now use $\epsilon = 10$. This change of $\epsilon$ is in DBSCAN's favour, to delay a very sharp decline in ARI for increasing dimensions observed with $\epsilon = 3$.

\begin{figure}[ht]
\centering
\subfigure{\includegraphics[width=0.42\textwidth]{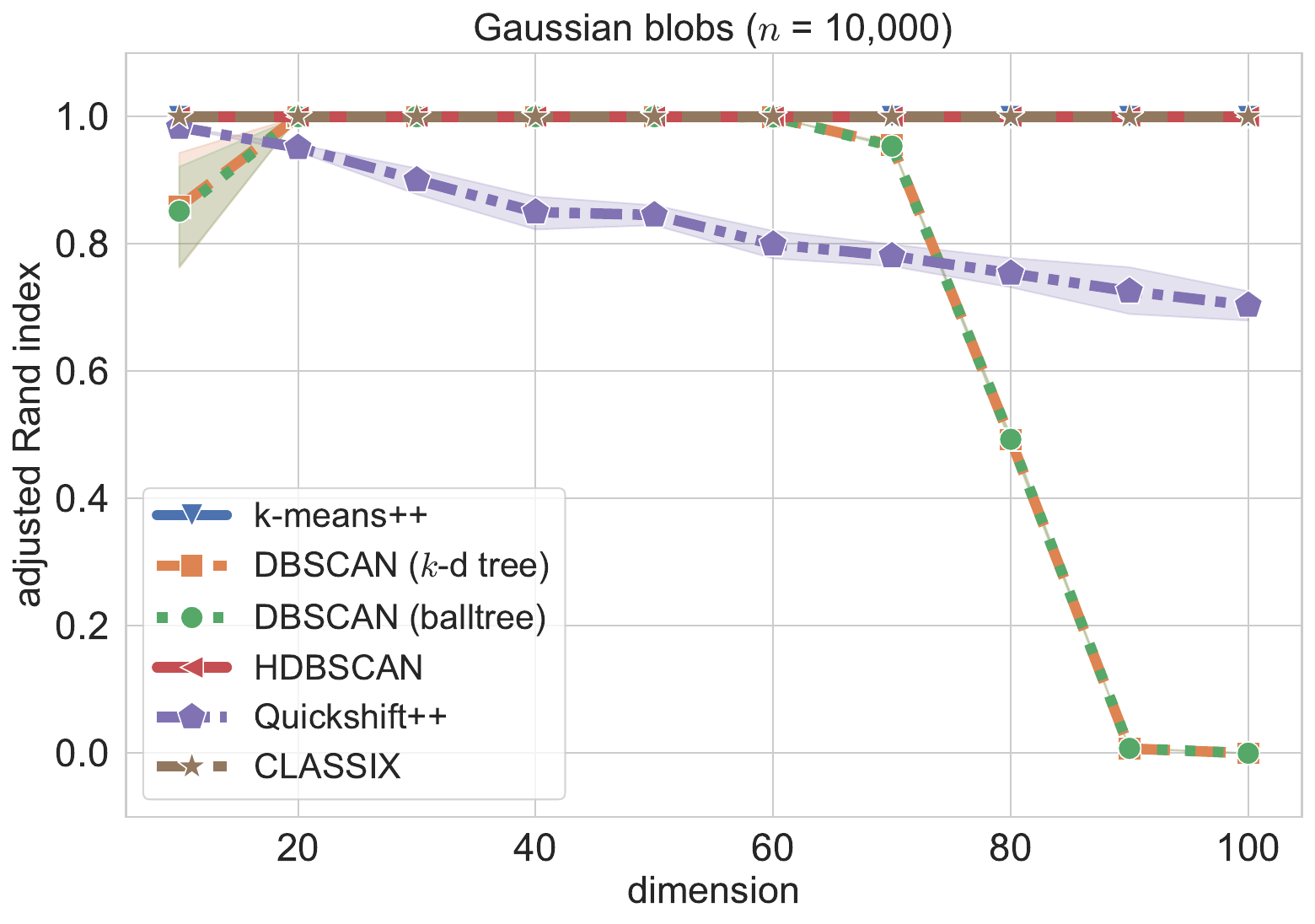}}
\ \ \subfigure{\includegraphics[width=0.41\textwidth]{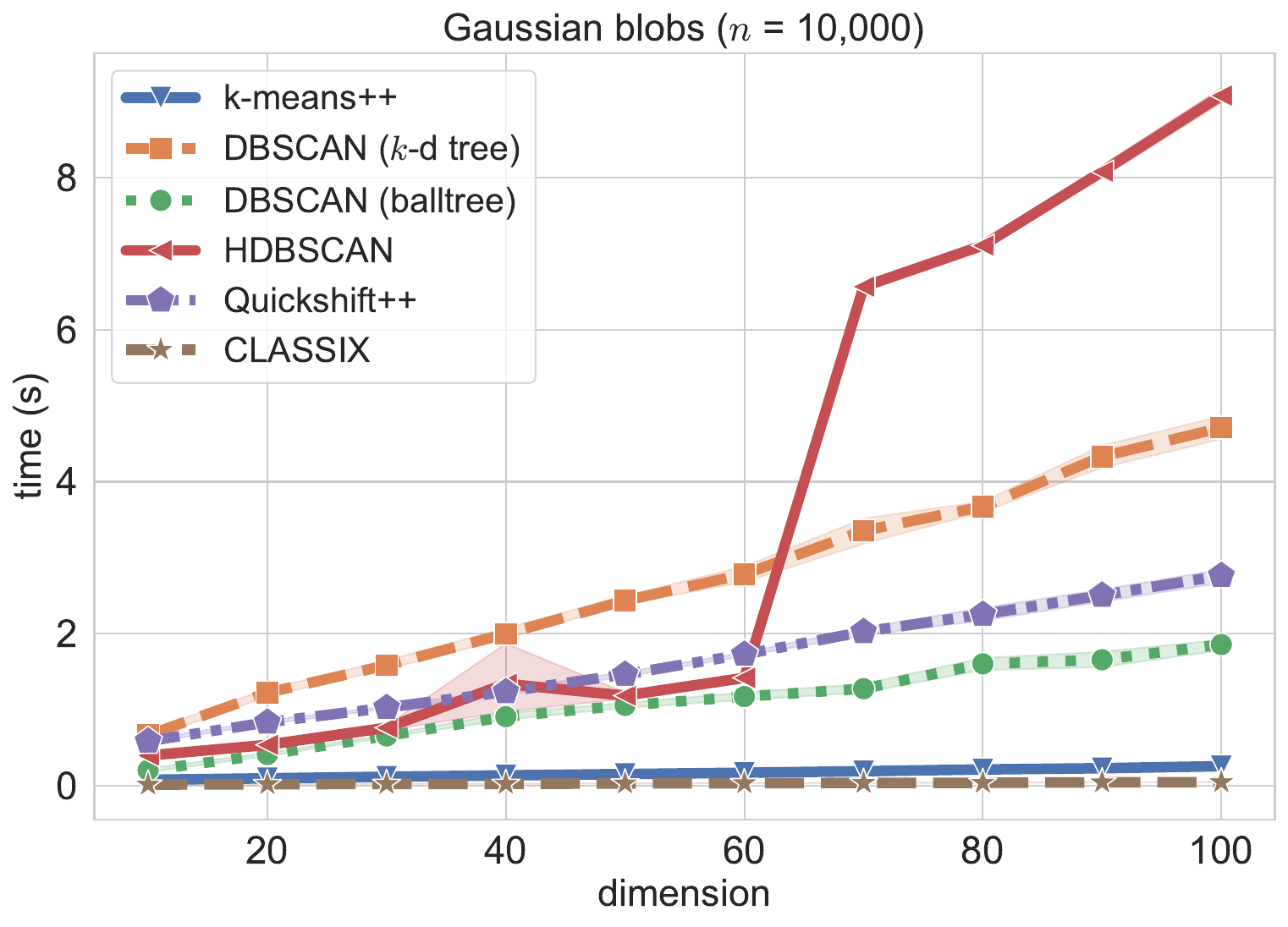}}\\
\centering\subfigure{\includegraphics[width=0.42\textwidth]{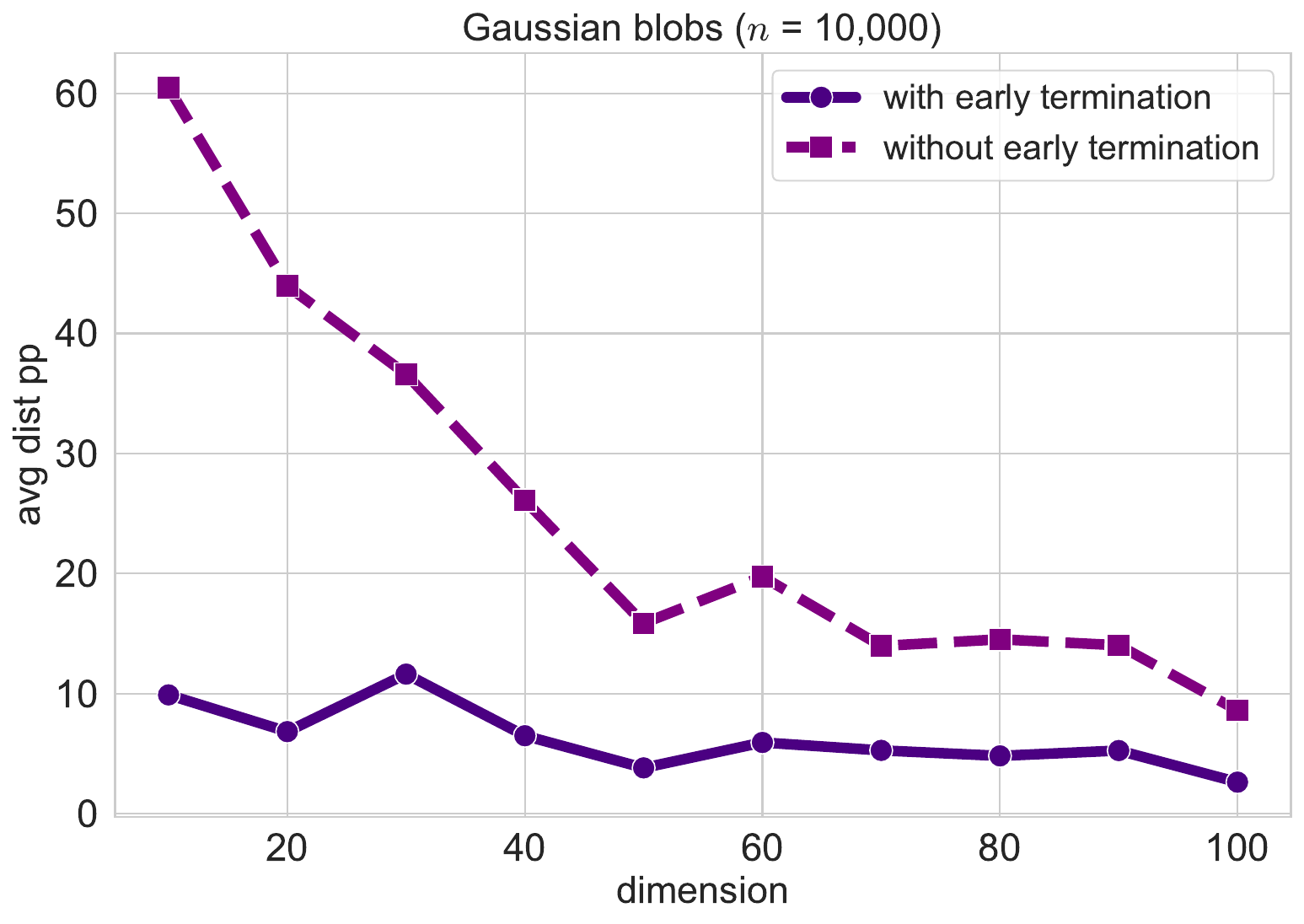}}
\caption[Scalability test]{Performance evaluation of various clustering methods on the Gaussian blobs data. Here we keep the number of  data points fixed at $n=10,000$ and vary the dimension from $d=10$ to 100. Top: The plots show the adjusted Rand index (left) and timing (right) for each method.
Bottom: Average number of distance calculations per data point in the CLASSIX aggregation phase with and without early termination conditions.}
\label{fig:size_dim_test}
\end{figure}

The results are shown \figurename~\ref{fig:size_dim_test}. In the plot on the top left we observe an ARI degradation of DBSCAN and Quickshift++ as $d$ increases. The hyperparameters of these two methods appear to be rather sensitive to the feature dimension. The other methods, including HDBSCAN and CLASSIX, perform robustly across all  dimensions without requiring hyperparameter adjustment. In terms of timings, we again find that \texttt{k-means++} and CLASSIX exhibit the best performance. For this data, the runtimes of all methods except HDBSCAN appear to scale linearly in the feature dimension~$d$. 

In the bottom of \figurename~\ref{fig:size_dim_test} we again show the average number of distance calculation per data point (\texttt{avg\_dist\_pp}) as the feature dimension~$d$ increases. We find that, both with and without early termination, \texttt{avg\_dist\_pp} decreases as $d$ increases. This is to be expected as a Gaussian blob with unit standard deviation becomes more ``focused'' in higher dimensions. Clustering a more focused point cloud with a fixed aggregation radius~$R$ is equivalent to clustering a constant point cloud with increasing radius~$R$. In line with the analysis of the Gaussian model in section~\ref{sec:simpmod}, we indeed expect the number of distance calculations to decrease in higher dimensions.

\subsubsection{Sensitivity to parameters}

We now focus our attention on the sensitivity of CLASSIX with respect to its  $\radius$ parameter.   \figurename~\ref{fig:sensitivity} shows the ARI and AMI scores of clustering with $\radius$ varying between 0.1 to 1.0. The data forms ten Gaussian blobs in $d=10$  dimensions and the number of data points~$n$ is varied from 1000 to 9000 in steps of 1000 (for each plot from the left to right and top to bottom). We show the ARI and AMI scores for both  distance-based and density-based CLASSIX clustering. We observe that the method is not very sensitive to the $\radius$ parameter. As long as this parameter is chosen within the interval 0.2 to 0.6, good to very good clustering results are obtained.

\newlength{\hcolww} 
\setlength{\hcolww}{0.30\textwidth}

\begin{figure}[ht]
	\centering
	\subfigure{\includegraphics[width=\hcolww]{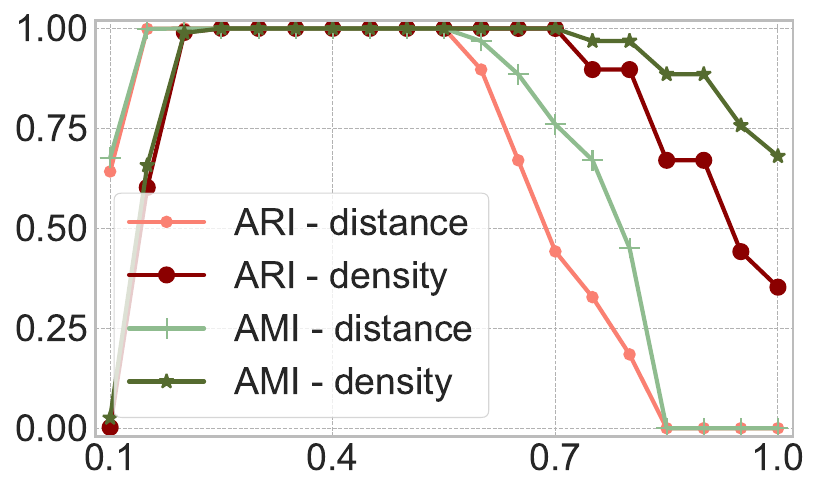}}
	\subfigure{\includegraphics[width=\hcolww]{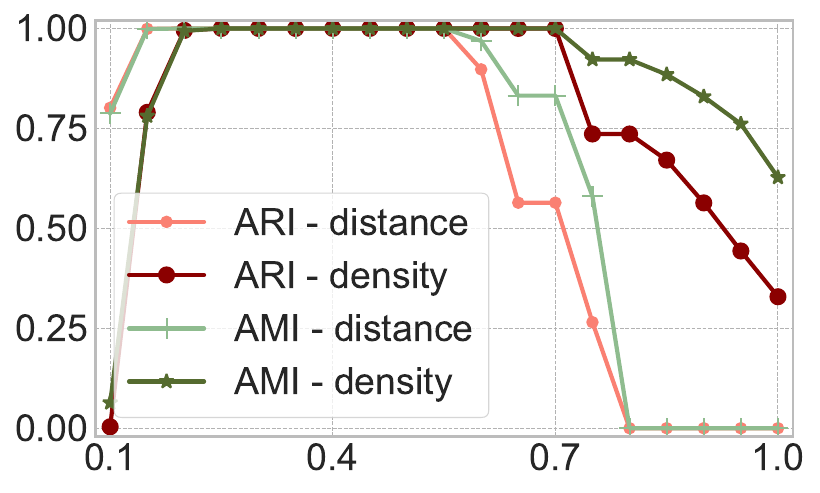}}
	\subfigure{\includegraphics[width=\hcolww]{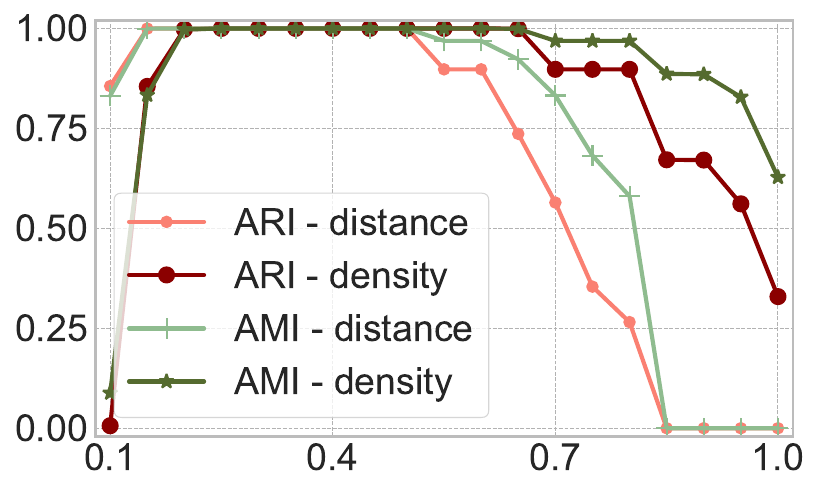}}
	\subfigure{\includegraphics[width=\hcolww]{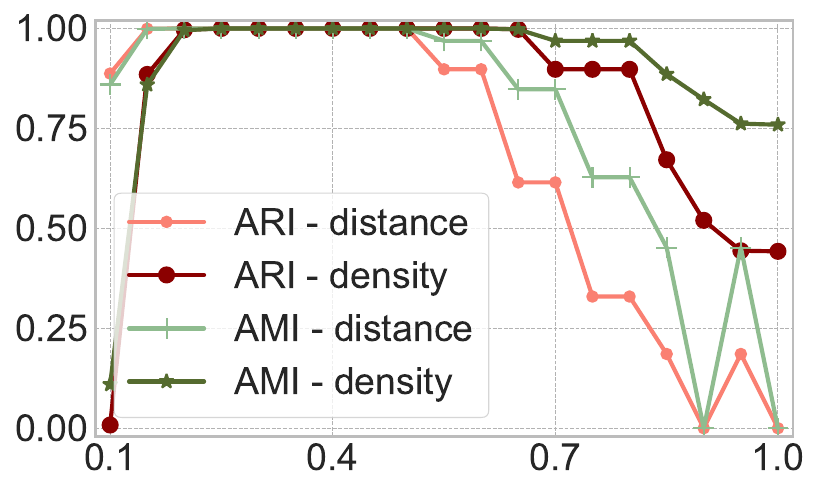}}
	\subfigure{\includegraphics[width=\hcolww]{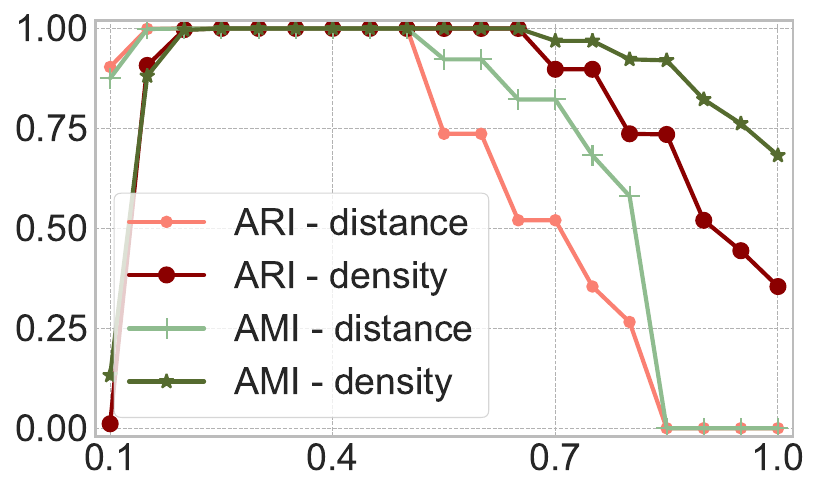}}
	\subfigure{\includegraphics[width=\hcolww]{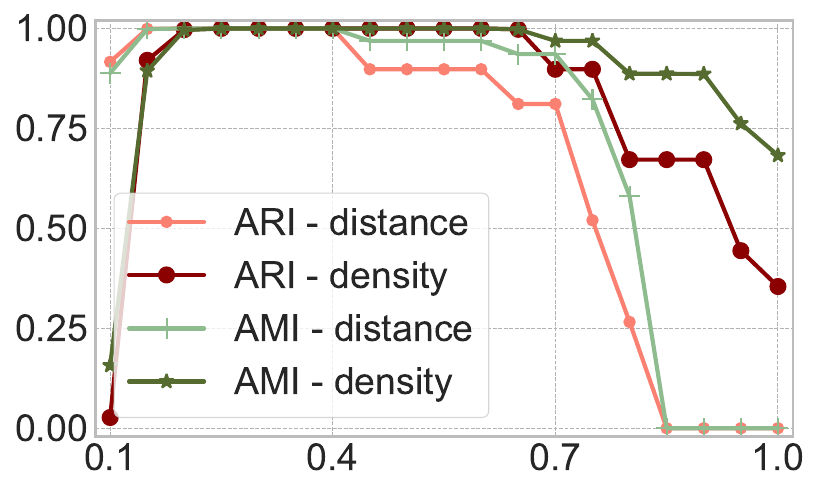}}
	\subfigure{\includegraphics[width=\hcolww]{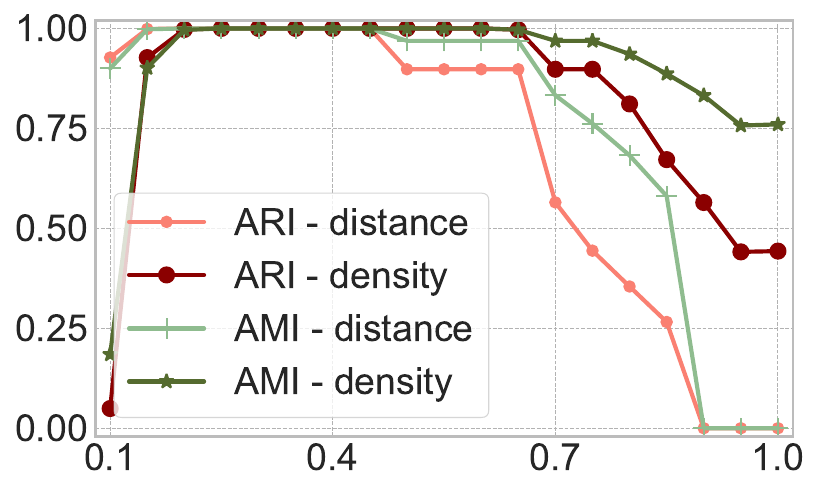}}
	\subfigure{\includegraphics[width=\hcolww]{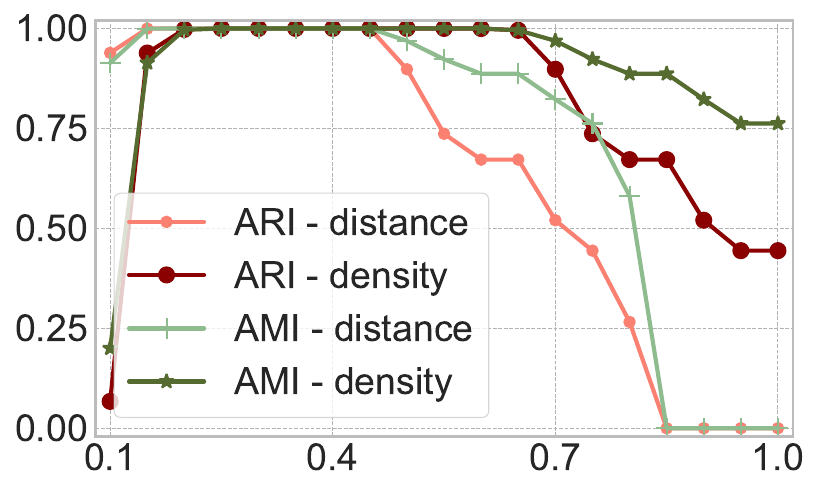}}
	\subfigure{\includegraphics[width=\hcolww]{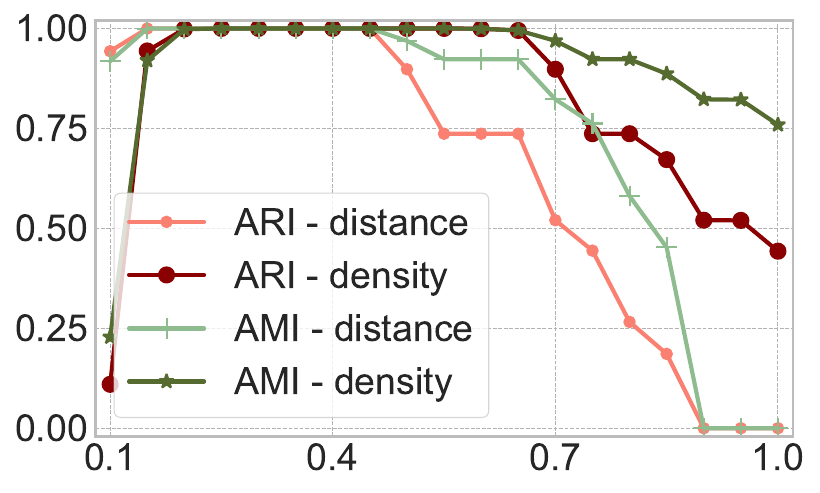}}
	
	\caption[Sensitivity to parameters]{Sensitivity of CLASSIX on the Gaussian blobs data as the  $\radius$ parameter is varied from $0.1$ to $1$. We report the ARI and AMI values for  distance-based and density-based merging.  CLASSIX produces  good clusterings over a wide parameter range from 0.2 to 0.6 (approximately).}
	\label{fig:sensitivity}
\end{figure}

\subsection{UCI Machine Learning Repository}
We now compare CLASSIX with DBSCAN, HDBSCAN, and Quickshift++  on a number of commonly-used real-world datasets listed in \tablename~\ref{realdata}. All of these datasets except ``Phoneme'' and ``Wine'' can be found 
in the UCI Machine Learning Repository~\cite{Dua:2019}.

\begin{table*}[ht]
\caption{Datasets in the UCI Machine Learning Repository}
\centering
\small
\setlength\tabcolsep{2.4pt}
\begin{tabular}{l c c c c}
    \toprule
    Dataset\ \ \  & Size & Dimension & Labels & Related references \\ [0.5ex] 
    \midrule
     Banknote & 1372 &	4 &	2 &	\cite{Dua:2019}\\
     \midrule
     Dermatology\ \ \ \  & 366 &	34 &	6 &	\cite{Dua:2019, Gvenir1998LearningDD}\\
      \midrule
     Ecoli & 336 &	7 &	8 &	\cite{Dua:2019, PMID:1946347, Nakai1992AKB}\\
      \midrule
     Glass & 214 &	9 & 6 & \cite{Dua:2019}\\
      \midrule
     Iris & 150 & 4 &	3 &	\cite{Dua:2019, FIS36, 10.2307/2394164}\\
     \midrule
     Phoneme & 4509 & 256 &	5 &	\cite{hastie_09_elements-of.statistical-learning}\\
      \midrule
     \multirow{2}{*}{Seeds} & \multirow{2}{*}{210} &	\multirow{2}{*}{7} & \multirow{2}{*}{3} &	\cite{Dua:2019}; \\
     &&&&\cite{10.1007/978-3-642-13105-9_2}\\
      \midrule
     Wine & 178 & 13 &	3 &	\cite{doi.org/10.1002/cem.1180040210}\\
 \bottomrule
\end{tabular}
\label{realdata}
\end{table*}

We preprocess the data by removing rows with missing values and z-normalise the features (i.e., shift each feature  to zero mean and scale it to unit variance). We then perform a hyperparameter search by measuring the ARI and AMI scores while varying the  main parameters of each method as shown in \figurename~\ref{fig:cluster_quality} and its caption. For each method and dataset  we use the respective best hyperparameters and report the achieved \rev{ARI,  AMI, FMI, and VM}  scores in \tablename~\ref{tab:real_data}. 
\rev{In terms of the average scores across all datasets shown on the bottom of  \tablename~\ref{tab:real_data}, CLASSIX achieves the best overall performance compared to the other methods,} in particular when density-based merging is used. The only exception is the ``Phoneme'' dataset where CLASSIX with distance-based merging significantly outperforms density-based merging. This might be explained by the rather high feature dimension of this dataset; see also the remark at the end of section~\ref{sec:dense}.

\newlength{\hcolw} 
\setlength{\hcolw}{0.16\textwidth}

\begin{figure}[ht]
	\centering
    \includegraphics[width=0.98\textwidth]{ARI_AMI_PARAMS.pdf}
	\caption[Clustering quality for datasets in the UCI Machine Learning Repository as the parameters change.]{Clustering quality on the UCI Machine Learning Repository  as the hyperparameters change. The rows correspond eight datasets in Table~\ref{realdata} while the columns correspond to Mean shift, DBSCAN,  HDBSCAN, Quickshift++,  CLASSIX with distance-based merging, and CLASSIX with density-based merging, respectively. The searched parameters  are bandwidth (Mean shift), $\epsilon$ (DBSCAN), minimum cluster size (HDBSCAN), $k$ (Quickshift++), and $\radius$ (CLASSIX).}
    \label{fig:cluster_quality}
\end{figure}

\begin{table*}[htbp]
\caption{\rev{Performance on datasets in the UCI Machine Learning Repository. Larger is better, with the respective maximal scores in each row highlighted.}   } 
\centering
\tiny
\setlength\tabcolsep{3.6pt}
\begin{tabular}{l c c c c c c c}
    \toprule
     &  & \multirow{2}{*}{Mean shift} & \multirow{2}{*}{DBSCAN}  & \multirow{2}{*}{HDBSCAN} & Quickshift & CLASSIX  & CLASSIX \\
     &  &   & & & ++ &(distance) & (density)\\
    \midrule
     \multirow{4}{*}{Banknote} & ARI &0.16&	0.80&	0.33&	0.34&	\best{0.87}&	0.85\\
     & AMI &0.25&	0.71&	0.42&	0.51&	\best{0.78} &	0.76\\
    & FMI  & 0.42	&0.90	&0.59	&0.59	&\best{0.93}	&\best{0.93}\\
    & VM & 0.25	& 0.71	& 0.42	& 0.51	& \best{0.78}	& 0.76\\

     \midrule
     \multirow{4}{*}{Dermatology}  & ARI &0.61&	0.28&	0.47&	0.65&	\best{0.68}&	\best{0.68}\\
     & AMI&0.77&	0.57&	0.66&	0.77&	\best{0.80}&	\best{0.80}\\
     & FMI & 0.73	& 0.57	& 0.60	& 0.74	& \best{0.75}	& \best{0.75}\\
     & VM & 0.78	& 0.58	& 0.67	& 0.78	& 0.80	& \best{0.81}\\

    \midrule
     \multirow{4}{*}{Ecoli}  & ARI &0.46&	0.50&	0.40&	\best{0.73}&	0.56&	0.67\\
     & AMI&0.44&	0.48&	0.41&	\best{0.68}&	0.58&	0.62\\
    & FMI  &0.67	&0.67	&0.59	&0.81	&0.72	&\best{0.76}\\
    & VM & 0.46 & 0.49 & 0.42 & \best{0.69} & 0.59	& 0.63\\

    \midrule
     \multirow{4}{*}{Glass}  & ARI &0.29&	0.25&	0.25&	\best{0.30}&	0.23&	0.28\\
     & AMI& \best{0.40}&	0.38&	0.37& \best{0.40}&	0.35&	0.38\\
    & FMI &0.55	&0.56	&0.56	&\best{0.57}	&0.55	&0.53\\
    & VM & 0.48	& 0.39	& 0.39	& 0.45	& 0.45	& \best{0.52}\\

    \midrule
     \multirow{4}{*}{Iris}  & ARI &0.57&	0.55&	0.56&	0.57&	0.56&	\best{0.83}\\
     & AMI&0.73&	0.68&	0.71&	0.73&	0.68&	\best{0.81}\\
    & FMI &0.77	&0.75	&0.76	&0.77	&0.73	&\best{0.89}\\
    & VM & 0.73	& 0.69	& 0.72	& 0.73	& 0.69	& \best{0.82}\\

    \midrule
     \multirow{4}{*}{Phoneme}  & ARI &0.41&	0.51&	0.41&	0.40&	\best{0.76}&	0.50\\
     & AMI &0.60&	0.66&	0.61&	0.65&	\best{0.85}&	0.57\\
    & FMI &0.61	&0.62	&0.41	&0.63	& \best{0.83}	&0.61\\
    & VM & 0.60	& 0.66	& 0.05	& 0.65	& \best{0.85}	& 0.57\\

    \midrule
     \multirow{4}{*}{Seeds}  & ARI &0.49&	0.40&	0.24&	\best{0.78}&	0.70&	0.71\\
     & AMI &0.56&	0.47&	0.40&	\best{0.72}&	0.68&	0.67\\
    & FMI &0.72	&0.62	&0.46	&\best{0.86}	&0.80	&0.80\\
    & VM & 0.56	& 0.47	& 0.42	& \best{0.72}	& 0.68	& 0.67\\

    \midrule
     \multirow{4}{*}{Wine}  & ARI &0.40&	0.44&	0.38&	0.76&	0.47&	\best{0.80}\\
     & AMI &0.47&	0.54&	0.46&	0.75&	0.61&	\best{0.76}\\
    & FMI &0.64	&0.65	&0.60	&0.84	&0.72	&\best{0.87}\\
    & VM & 0.50	& 0.55	& 0.47	& 0.75	& 0.61	& \best{0.76}\\

     \midrule\midrule
    \multirow{4}{*}{Average}  & ARI &0.42&	0.47&	0.38&	0.57&	0.60&	\best{0.66}\\
     & AMI &0.53&	0.56&	0.51&	0.65&	\best{0.67}&	\best{0.67}\\
    & FMI &0.64	&0.67	&0.57	&0.73	&0.76	&\best{0.77}\\
    & VM &0.55	& 0.57	& 0.44	& 0.66	& 0.68	& \best{0.69}\\
    \bottomrule
\end{tabular}
\label{tab:real_data}
\end{table*}

\subsection{Scikit-learn benchmark}\label{scikit_learn_benchmark}\rev{The scikit-learn library contains a  benchmark\footnote{\url{https://scikit-learn.org/stable/modules/clustering.html}} with six datasets and a number of clustering methods such as BIRCH \cite{10.1145/235968.233324},  spectral clustering \cite{Shi00normalizedCuts, yu2003multiclass},  and Gaussian mixture models (GMM) \cite[Chap.~11]{Murphy2012}}. The library provides preselected parameters for all of these methods and we have left them unchanged. We added HDBSCAN, Quickshift++, \rev{dePDDP\footnote{Implementation available from~\url{https://github.com/panagiotisanagnostou/HiPart}}, and CLASSIX with distance-based and density-based merging to the benchmark. We provide the true number of clusters to \texttt{k-means++} and dePDDP (as the \texttt{max\_clusters\_number} parameter) and leave the other parameters at their default values. The other clustering methods are tuned by grid search to find the best hyperparameters in terms of the achieved ARI and AMI score.}

\begin{figure*}[ht]
	\centering
	\includegraphics[width=\textwidth]{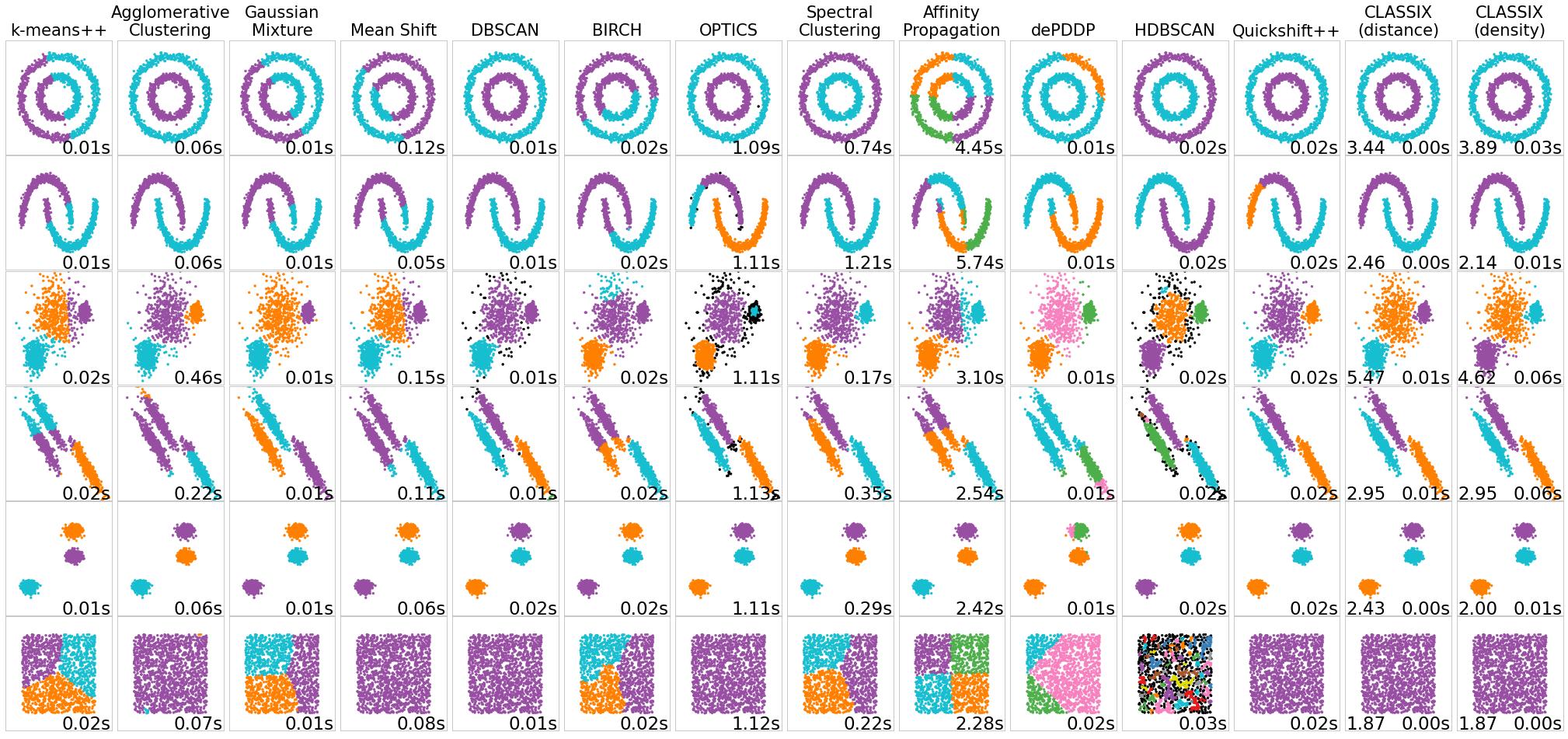}
	\caption[Clustering comparison on the scikit-learn benchmark]{Visualization of clustering results on the scikit-learn benchmark. Each plot shows the required computation time on the bottom right, and for CLASSIX the average number of distance computations per data point on the bottom left. }
	\label{Vis-toydatasets}
\end{figure*}

The  clusterings computed by each method are visualized in \figurename~\ref{Vis-toydatasets}. On all five datasets, CLASSIX produces very good clusters. Among the other methods, Quickshift++ stands out with comparable performance. The figure also shows the required computation time in the bottom right of each plot, and for CLASSIX the average number of distance computations per data point in the bottom left. Most methods can be considered fast, with just a few milliseconds of computation time required, but CLASSIX consistently performs fastest here. This is explained by the small number of distance computations that are required, on average never exceeding 5.47 comparisons per data point in these five tests. 
\rev{The four clustering quality metrics}  are shown in \tablename~\ref{tab:toydatasets}. Both Quickshift++ and CLASSIX stand out as the best performing methods here, \rev{with CLASSIX (distance) outperforming all other methods across all four quality metrics on average, closely followed by CLASSIX (density).}

\begin{table*}[htbp]
\caption{\rev{Performance on the six datasets in the scikit-learn clustering  benchmark. Larger is better, with the respective maximal scores in each row highlighted. The last four rows ``AVG'' correspond to the averages across all datasets.  }} 
\tiny
\centering
\setlength\tabcolsep{3.7pt}
\begin{tabular}{c c c c c c c c c c c c c c c c}
\toprule
\multirow{2}{*}{}  &  & \multirow{2}{*}{\texttt{k-means++}}  & Aggl. & \multirow{2}{*}{GMM} &  {Mean}  &  \multirow{2}{*}{DBSCAN} &  \multirow{2}{*}{BIRCH}  & \multirow{2}{*}{OPTICS}&  Spectral &Affinity& \multirow{2}{*}{dePDDP} & \multirow{2}{*}{HDBSCAN} & Quickshift & CLASSIX & CLASSIX \\
& & & clust.&     & shift & & & & clust. & prop. & & & ++ & (distance) & (density)\\
\midrule
\multirow{4}{*}{I} & ARI & 0.00&	\best{1.00}&	0.00&	0.01&	\best{1.00}&	0.00&	\best{1.00}&	\best{1.00}&	0.00&	0.06&	\best{1.00}&	\best{1.00}&	\best{1.00}&	\best{1.00} \\
& AMI &0.00&	\best{1.00}&	0.00&	0.01&	\best{1.00}&	0.00&	\best{1.00}&	\best{1.00}&	0.00&	0.17&	\best{1.00}&	\best{1.00}&	\best{1.00}&	\best{1.00}\\
& FMI &0.50	& \best{1.00}&	0.50&	0.53&	\best{1.00}&	0.51&	\best{1.00}&	\best{1.00}&	0.35&	0.65&	\best{1.00}&	\best{1.00}&	\best{1.00}&	\best{1.00}\\
& VM &0.00&	\best{1.00}&	0.00&	0.01&	\best{1.00}&	0.00&	\best{1.00}&	\best{1.00}&	0.00&	0.17&	\best{1.00}&	\best{1.00}&	\best{1.00}&	\best{1.00}\\
\midrule
\multirow{4}{*}{II} & ARI &0.48&	\best{1.00}&	0.50&	0.54&	\best{1.00}&	0.57&	0.77&	\best{1.00}&	0.34&	0.47&	\best{1.00}&	0.78&	\best{1.00}&	\best{1.00}\\
& AMI &0.38&	\best{1.00}&	0.40&	0.44&	\best{1.00}&	0.56&	0.79&	\best{1.00}&	0.37&	0.39&	\best{1.00}&	0.81&	\best{1.00}&	\best{1.00}\\
& FMI &0.74&	\best{1.00}&	0.75&	0.77&	\best{1.00}&	0.79&	0.88&	\best{1.00}&	0.59&	0.74&	\best{1.00}&	0.88&	\best{1.00}&	\best{1.00}\\
& VM &0.38&	\best{1.00}&	0.40&	0.44&	\best{1.00}&	0.56&	0.79&	\best{1.00}&	0.37&	0.39&	\best{1.00}&	0.81&	\best{1.00}&	\best{1.00}\\

\midrule
\multirow{4}{*}{III} & ARI &0.81&	0.93&	\best{0.97}&	0.84&	0.55&	0.55&	0.71&	0.94&	0.81&	0.94&	0.86&	0.94&	0.95&	0.92\\
& AMI&0.80&	0.90&	\best{0.94}&	0.83&	0.66&	0.66&	0.73&	0.91&	0.80&	0.91&	0.84&	0.92&	0.92&	0.89\\
& FMI &0.87&	0.95&	\best{0.98}&	0.90&	0.75&	0.74&	0.80&	0.96&	0.88&	0.96&	0.90&	0.96&	0.97&	0.95\\
& VM &0.80&	0.90&	\best{0.94}&	0.83&	0.66&	0.66&	0.73&	0.91&	0.80&	0.91&	0.84&	0.92&	0.92&	0.89\\
	
\midrule
\multirow{4}{*}{IV} & ARI &0.61&	0.42&	\best{1.00}&	0.53&	0.97&	0.57&	0.95&	0.96&	0.62&	0.42&	0.89&	\best{1.00}&	\best{1.00}&	\best{1.00}\\
& AMI&0.62&	0.53&	\best{1.00}&	0.63&	0.95&	0.63&	0.92&	0.94&	0.62&	0.55&	0.86&	\best{1.00}&	\best{1.00}&	\best{1.00}\\
& FMI &0.74&	0.69&	\best{1.00}&	0.75&	0.98&	0.72&	0.96&	0.97&	0.74&	0.68&	0.93&	\best{1.00}&	\best{1.00}&	\best{1.00}\\
& VM &0.62&	0.53&	\best{1.00}&	0.63&	0.96&	0.63&	0.92&	0.94&	0.62&	0.55&	0.86&	\best{1.00}&	\best{1.00}&	\best{1.00}\\

\midrule
\multirow{4}{*}{V} & ARI &\best{1.00} &	\best{1.00} &	\best{1.00} &	\best{1.00} &	\best{1.00} &	\best{1.00} &	\best{1.00} &	\best{1.00} &	\best{1.00} &	0.49 &	\best{1.00} &	\best{1.00} &	\best{1.00}&	\best{1.00}\\
& AMI&\best{1.00}&	\best{1.00}&	\best{1.00}&	\best{1.00}&	\best{1.00}&	\best{1.00}&	\best{1.00}&	\best{1.00}&	\best{1.00}&	0.66&	\best{1.00}&	\best{1.00}&	\best{1.00}&	\best{1.00}\\
& FMI &\best{1.00}&	\best{1.00}&	\best{1.00}&	\best{1.00}&	\best{1.00}&	\best{1.00}&	\best{1.00}&	\best{1.00}&	\best{1.00}&	0.72&	\best{1.00}&	\best{1.00}&	\best{1.00}&	\best{1.00}\\
& VM &\best{1.00}&	\best{1.00}&	\best{1.00}&	\best{1.00}&	\best{1.00}&	\best{1.00}&	\best{1.00}&	\best{1.00}&	\best{1.00}&	0.66&	\best{1.00}&	\best{1.00}&	\best{1.00}&	\best{1.00}
\\

\midrule
\multirow{4}{*}{VI} & ARI &0.00&	0.00&	0.00&	\best{1.00}&	\best{1.00}&	0.00&	\best{1.00}&	0.00&	0.00&	0.00&	0.00&	\best{1.00}&	\best{1.00}&	\best{1.00}\\
& AMI&0.00&	0.00&	0.00&	\best{1.00}&	\best{1.00}&	0.00&	\best{1.00}&	0.00&	0.00&	0.00&	0.00&	\best{1.00}&	\best{1.00}&	\best{1.00}\\
& FMI &0.58&	0.99&	0.58&	\best{1.00}&	\best{1.00}&	0.59&	\best{1.00}&	0.58&	0.50&	0.74&	0.42&	\best{1.00}&	\best{1.00}&	\best{1.00}\\
& VM &0.00&	0.00&	0.00&	\best{1.00}&	\best{1.00}&	0.00&	\best{1.00}&	0.00&	0.00&	0.00&	0.00&	\best{1.00}&	\best{1.00}&	\best{1.00}
\\

\midrule \midrule
\multirow{4}{*}{AVG} & ARI &0.48&	0.72&	0.58&	0.65&	0.92&	0.45&	0.90&	0.82&	0.46&	0.40&	0.79&	0.95&	\best{0.99}&	\best{0.99}\\
& AMI& 0.47&	0.74&	0.56&	0.65&	0.94&	0.47&	0.91&	0.81&	0.47&	0.45&	0.78&	0.95&	\best{0.99}&	0.98\\
& FMI &0.74&	0.94&	0.80&	0.82&	0.96&	0.72&	0.94&	0.92&	0.68&	0.75&	0.87&	0.97&	\best{0.99}&	\best{0.99}\\
& VM &0.47&	0.74&	0.56&	0.65&	0.94&	0.47&	0.91&	0.81&	0.47&	0.45&	0.78&	0.96&	\best{0.99}
&	0.98\\
\bottomrule
\end{tabular}
\label{tab:toydatasets}
\end{table*}

\subsection{Shape clusters test}\label{shape_benchmark}
This benchmark involves eight synthetic datasets with varying shapes that are challenging for clustering algorithms. The properties of each dataset are detailed in \tablename~\ref{table:shapeset}. Before clustering this data, we z-normalise each of the features. 

\begin{table*}[ht]
	\caption{Datasets in the Shape benchmark used for our tests} 
	\centering 
	\small
	\begin{tabular}{c l c c c c c c c c c c} 
		\toprule 
	 Id & Dataset &  Size & Labels  & Related references\\ [0.5ex] 
        \midrule
		(i) & Aggregation &  788 &  7 &  \cite{Aggregation} \\ 
		(ii) & Compound &  399 &  6 & \cite{Compound} \\ 
        (iii) & D31 & 3100 & 31 & \cite{D31R15} \\ 
		(iv) & Flame &  240 &  2 &   \cite{Flame} \\ 
		(v) & Jain &  373 &  2 & \cite{Jain} \\ 
	    (vi) & Pathbased &  300 & 3 & \cite{PathbasedSpiral} \\ 
		(vii) & R15 &  600 & 15 &  \cite{D31R15} \\ 		
		(viii) & Spiral &  312 & 3 & \cite{PathbasedSpiral} \\ 
		\bottomrule
	\label{table:shapeset}
	\end{tabular}
\end{table*}

\begin{figure*}[thbp]
	\centering
    \includegraphics[width=0.9\textwidth]{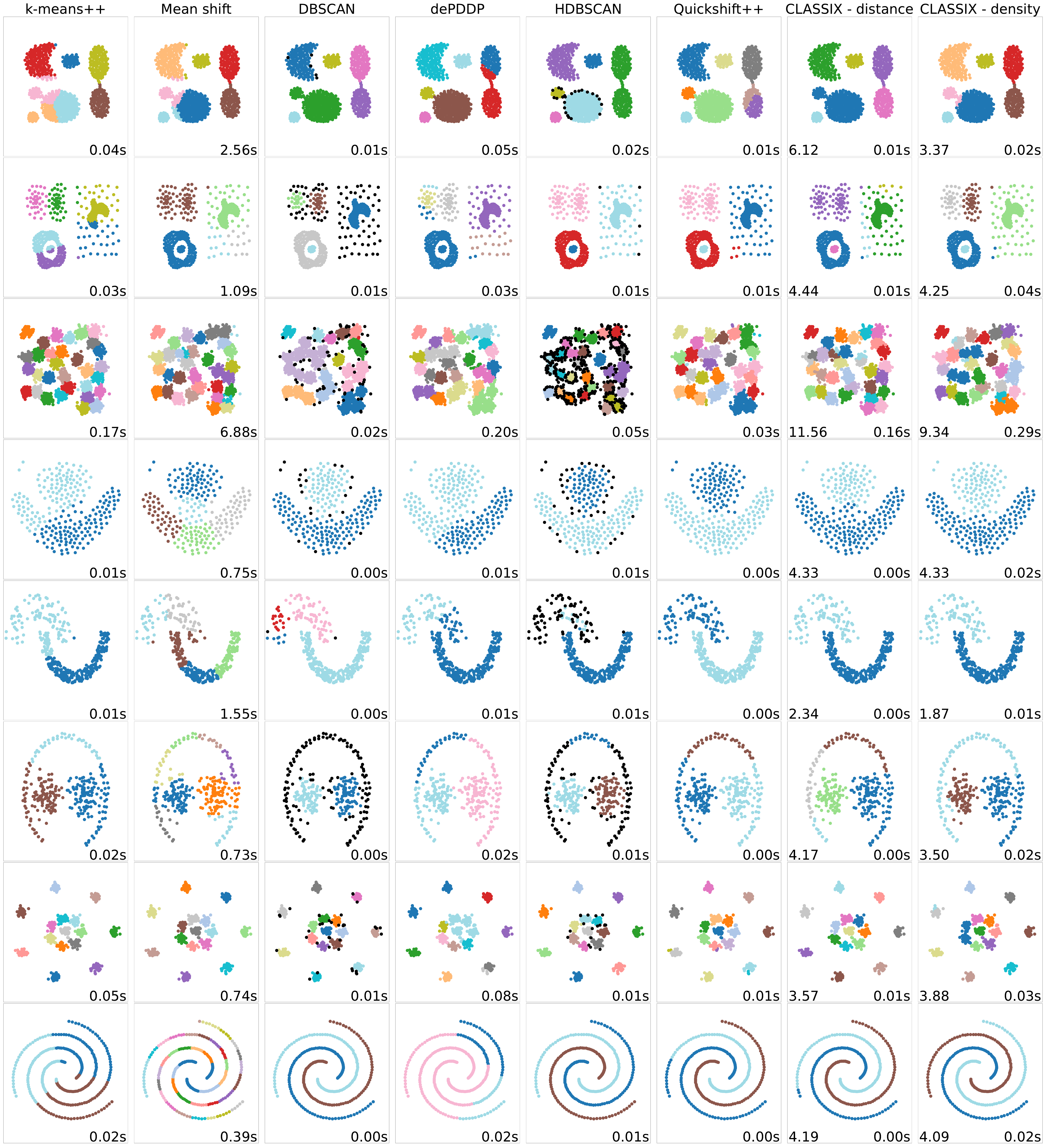}
	\caption{Visualisation of clustering results on the Shape benchmark. Each plot shows the required computation time on the bottom right, and for CLASSIX the average number of distance computations per data point on the bottom left.}
	\label{vis-Shape}
\end{figure*}

We compare \texttt{k-means++}, Mean shift, DBSCAN, dePDDP, HDBSCAN, Quickshift++, CLASSIX (distance-based), and CLASSIX (density-based). For \texttt{k-means++} \rev{and dePDDP} we specify the ground-truth number~$k$, while the hyperparameters of the other methods are again optimized by grid search. 
The clustering results are visualized in \figurename~\ref{vis-Shape}, together with the required computation time (bottom right of each plot) and for CLASSIX, the average number of distance computations per data point shown in the bottom left of the CLASSIX plots. We find that CLASSIX with density-based merging produces slightly better clusters than distance-based merging,  but at the expense of a  higher computation time. 

The measured \rev{AMI, ARI, FMI, and VM} scores are listed in \tablename~\ref{tab:shape_data}. Here we see a  more mixed picture, with algorithms performing rather differently depending on the dataset. Clearly, in view of \figurename~\ref{vis-Shape}, \rev{cluster quality metrics like AMI or ARI} do not always provide a good measure of what would be considered a ``good clustering'' in human perception. Still, on average, CLASSIX with density-based merging \rev{performs best across all datasets, closely followed by CLASSIX (distance-based),  Quickshift++, DBSCAN, and HDBSCAN (in that order).}


\begin{table*}[htbp]
\caption{\rev{Performance of various clustering algorithms on the Shape benchmark. Larger is better, with the respective maximal scores in each row highlighted.}} 
\centering
\tiny
\setlength\tabcolsep{6pt}
\begin{tabular}{c c c c c c c c c c}
    \toprule
       & & \multirow{2}{*}{k-means++} &	Mean &	\multirow{2}{*}{DBSCAN} &
    \multirow{2}{*}{dePDDP} &
    \multirow{2}{*}{HDBSCAN} &	Quickshift &	CLASSIX  &	CLASSIX \\
     & & &	 shift &	 &&	 &++ & (distance) &	(density)\\
    \midrule
     \multirow{4}{*}{Aggregation}& ARI & 0.72&	0.86&	0.90&	0.93&	0.84&	\best{0.97} &	0.92&	0.96\\
     & AMI & 0.83&	0.89&	0.94&	0.93&	0.90&	\best{0.97}&	0.96&	\best{0.97}\\
     &FMI &0.78&	0.89&	0.93&	0.94&	0.87&	\best{0.98}&	0.94&	0.97\\
     &VM &0.83&	0.89&	0.94&	0.93&	0.90&	\best{0.97}&	0.96&	\best{0.97}\\
     \midrule
     \multirow{4}{*}{Compound} & ARI & 0.57&	0.77&	\best{0.92}&	0.75&	0.81&	0.79&	0.82&	0.85\\
     & AMI &0.72&	0.79&	\best{0.89} &	0.77&	0.85&	0.84&	0.83&	\best{0.89} \\
     &FMI &0.67&	0.84&	\best{0.94} &	0.83&	0.87&	0.86&	0.87&	0.89\\
     &VM&0.72&	0.79&	\best{0.90}&	0.77&	0.86&	0.84&	0.84&	0.89\\
    \midrule
     \multirow{4}{*}{D31} & ARI & \best{0.96}&	0.94&	0.31&	0.84&	0.60&	0.93&	0.90&	0.83\\
     & AMI &\best{0.97}&	0.95&	0.74&	0.93&	0.84&	0.95&	0.94&	0.93\\
     &FMI &\best{0.96}&	0.94&	0.45&	0.85&	0.62&	0.93&	0.90&	0.84\\
     &VM&\best{0.97}&	0.96&	0.74&	0.94&	0.85&	0.95&	0.94&	0.93\\
     
    \midrule
     \multirow{4}{*}{Flame} & ARI & 0.42&	0.36&	0.84&	0.30&	0.73&	\best{1.00}&	0.87&	0.97\\
     &AMI &0.38&	0.50&	0.75&	0.37&	0.66&	\best{1.00}&	0.81&	0.94\\
     &FMI&0.72&	0.62&	0.92&	0.67&	0.86&	\best{1.00}&	0.94&	0.98\\
     &VM&0.39&	0.51&	0.75&	0.37&	0.66&	\best{1.00}&	0.81&	0.94\\

    \midrule
     \multirow{4}{*}{Jain} & ARI & 0.55&	0.27&	0.94&	0.62&	0.92&	\best{1.00}&	\best{1.00}&	\best{1.00}\\
     &AMI &0.51&	0.45&	0.84&	0.55&	0.82&	\best{1.00}&	\best{1.00}&	\best{1.00}\\
     &FMI&0.81&	0.57&	0.98&	0.88&	0.97&	\best{1.00}&	\best{1.00}&	\best{1.00}\\
     &VM&0.51&	0.46&	0.84&	0.55&	0.82&	\best{1.00}&	\best{1.00}&	\best{1.00}\\

    \midrule
     \multirow{4}{*}{Pathbased} & ARI & 0.48&	0.52&	\best{0.75}&	0.46&	0.67&	0.47&	0.61&	0.68\\
     &AMI &0.56&	0.52&	\best{0.74}&	0.53&	0.66&	0.55&	0.70&	0.73\\
     &FMI&0.67&	 0.66&	\best{0.83}&	0.66&	0.77&	0.67&	0.74&	0.79\\
     &VM&0.56&	0.53&	\best{0.74}&	0.53&	0.67&	0.55&	0.70&	0.73\\
    \midrule
     \multirow{2}{*}{R15} & ARI & \best{0.99}&	\best{0.99}&	0.90&	0.78&	0.95&	0.98&	0.98&	0.91\\
     &AMI &\best{0.99}&	\best{0.99}&	0.92&	0.93&	0.96&	0.98&	\best{0.99}&	0.97\\
     &FMI&\best{0.99}&	\best{0.99}&	0.91&	0.80&	0.95&	0.98&	0.98&	0.92\\
     &VM&\best{0.99}&	\best{0.99}&	0.93&	0.93&	0.96&	\best{0.99}&	\best{0.99}&	0.97\\
    \midrule
     \multirow{4}{*}{Spiral} & ARI &  -0.01&	0.12&	\best{1.00}&	0.04&	\best{1.00}&	0.88&	0.97&	\best{1.00}\\
     &AMI &-0.01&	0.43&	\best{1.00}&	0.12&	\best{1.00}&	0.88&	0.96&	\best{1.00}\\
     &FMI&0.33&	0.30&	\best{1.00}&	0.46&	\best{1.00}&	0.92&	0.98&	\best{1.00}\\
    &VM&0.00&	0.46&	\best{1.00}&	0.13&	\best{1.00}&	0.88&	0.96&	\best{1.00}\\
     \midrule \midrule
    \multirow{4}{*}{Average} & ARI &  0.59&	0.60&	0.82&	0.59&	0.81&	0.88&	0.88&	\best{0.90}\\
     &AMI & 0.62& 0.69&	0.85&	0.64&	0.84&	0.90&	0.90&	\best{0.93}\\
     &FMI&0.74&	0.73&	0.87&	0.76&	0.87&	\best{0.92}&	\best{0.92}&	\best{0.92}\\
     &VM&0.62&	0.70&	0.86&	0.64&	0.84&	0.90&	0.90&	\best{0.93}\\
    \bottomrule
\end{tabular}
\label{tab:shape_data}
\end{table*}

\subsection{Facial image classification}
Most often, image classification is done by supervised learning techniques such as deep convolutional neural networks \cite{NIPS2012_c399862d, Simonyan14a} instead of unsupervised learning techniques. Nevertheless, there can be benefit in using  unsupervised clustering algorithms in situations where labels are not provided. 

\begin{figure*}[ht]
	\centering
	\includegraphics[width=0.75\textwidth]{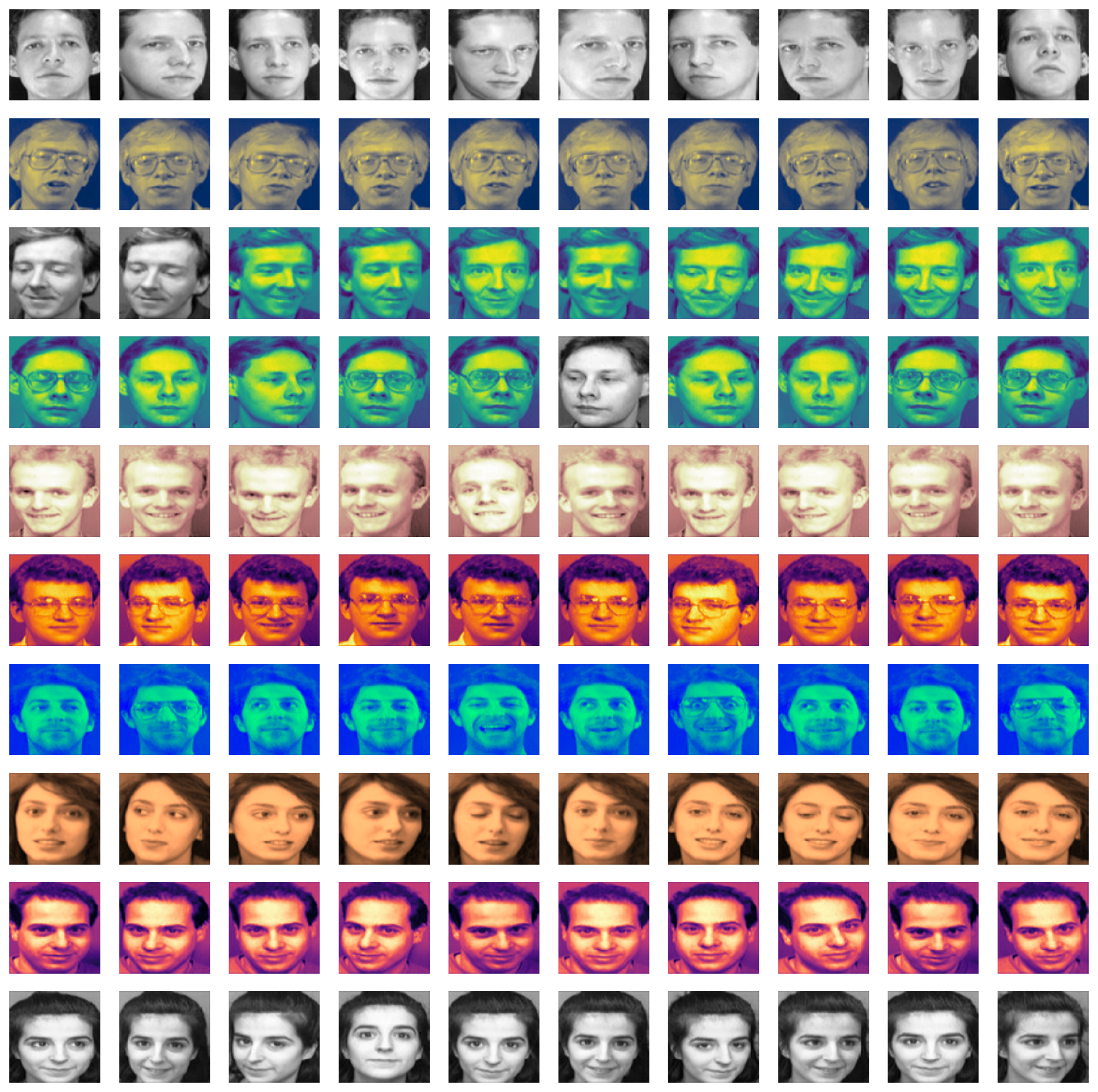}
	\caption[Facial clustering]{CLASSIX clustering performance on the Olivetti face dataset. Different colours indicate different clusters. The gray-scale images are outliers that have not been assigned to any cluster. The computation time is about 0.33s.}
	\label{faces}
\end{figure*}

 We apply CLASSIX to classify  images in the Olivetti face dataset. 
 The dataset contains 400 images of 40 distinct faces with variations in lighting, facial expressions (e.g., smiling or not), and other facial details (e.g., wearing glasses or not). 
Following the example in \cite{Rodriguez1492}, we select ten variants of ten faces from the dataset. The images are resized to $100\times 100$ pixels and  represented as $d=100^2$-dimensional feature vectors containing the gray-scale values. We  run distance-based CLASSIX with $\radius=0.6$ and $\minPts=7$ to cluster these faces into groups of similar images, hopefully corresponding to the same person. The results are shown in \figurename~\ref{faces} with different colours corresponding to different clusters. Images shown in gray-scale are outliers, i.e., not assigned to any cluster. 

We see from \figurename~\ref{faces}  that CLASSIX classifies 77 of the 100~images into 7~clusters. Six of these clusters contain ten images corresponding to the correct classes, i.e., these 60 images (60\% of all) are perfectly classified. The seventh cluster mixes 8~images of one face with 9~images of another similarly looking face. We may distinguish two types of misclassification as follows: type~(A) where images of a unique face are assigned to different clusters, and type~(B) where images showing different faces are assigned to the same cluster. Using these types, CLASSIX produces no type~(A) misclassifications on this data, but two faces have been type~(B) misclassified. 

For comparison, the method by \cite{Rodriguez1492} classifies 41 out of the 100 images into 9~clusters. Of these, 33~images (33\% of all) are perfectly classified into 7~clusters. The remaining 8~images correspond to the same face and have been type~(B) misclassified into two clusters. There are no type~(A) misclassifications. 


\subsection{Image segmentation}
\begin{figure*}[ht]
	\centering
	\includegraphics[width=0.8\textwidth]{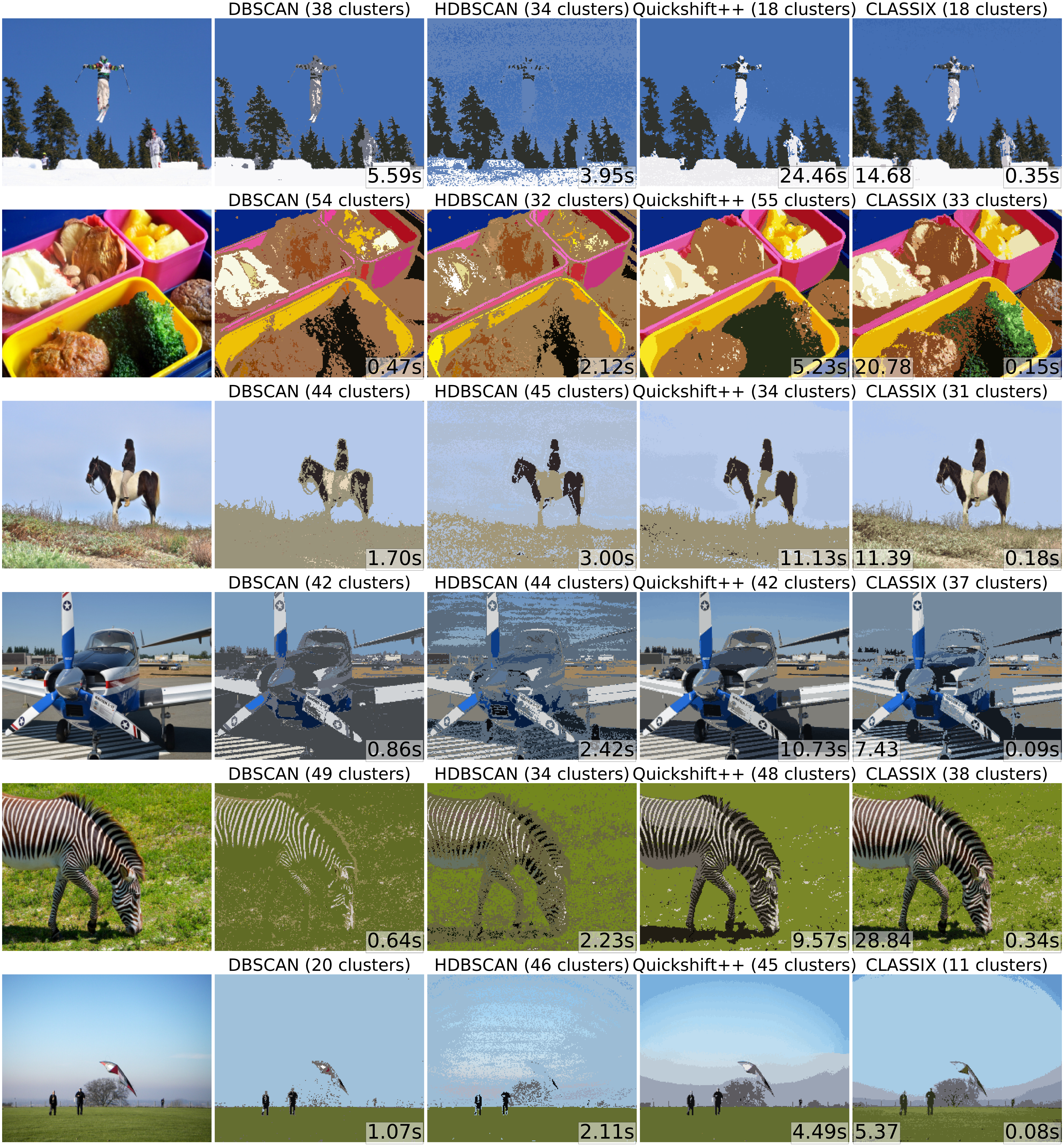}
	\caption[Density clustering on image]{Image segmentation via clustering. The left-most column shows the original image. For each method and image, the required clustering time is shown on the bottom right. For CLASSIX, the average number of distance calculations per data point is shown on the bottom left.}
	\label{fig:imgsegment}
\end{figure*}

Image segmentation is a technique to partition an image into  regions sharing  similar characteristics, usually for the purpose of identifying objects. Here we demonstrate an application of CLASSIX to cluster images on a pixel-by-pixel level. 
To this end we select five images from the COCO 2017 dataset \cite{10.1007/978-3-319-10602-1_48}, as shown in \figurename~\ref{fig:imgsegment}. Each RGB image of size 100-by-100 pixels is converted into a data array of shape $[100^2, 3]$, and we then cluster these $n=100^2$ data points of dimension $d=3$ using DBSCAN, HDBSCAN, Quickshift++, and CLASSIX with distance-based merging. After the clusters are formed, each pixel feature is replaced by the mean of all points in the cluster it is associated with, resulting in the segmented image. 

It is very challenging to choose the parameters of all these methods so that the number of resulting clusters is fairly comparable. Here we have used $\radius=0.15$ and $\texttt{scale}=1.05$ for CLASSIX 
and then manually tuned \texttt{minPts} and the hyperparameters of the other clustering algorithms to produce  a comparable, preferably slightly higher number of clusters. As can be seen in \figurename~\ref{fig:imgsegment}, CLASSIX is by far the fasted method in this test, producing  reconstructions of high visual quality compared to the other methods despite using the smallest number of clusters. The most challenging image (Zebra) required about 0.33s to segment with an average number of distance calculations per data point of~28.84.

\section{Summary and future work}\label{Section: Conclusion}

We have introduced a fast  clustering algorithm based on the sorting of the data points by their first principal coordinate. CLASSIX's key component is the fast aggregation of nearby data points into groups, exploiting the sorted order to reduce the number of pairwise distance computations. The simplicity of the aggregation and merging phases allows for clustering results to be explainable, a property that we demonstrate in the Appendix. 

We have tested the clustering efficiency of CLASSIX in a  number of experiments, ranging from low to high-dimensional data. \rev{We found the method to be competitive in terms of speed and cluster quality, and relatively easy to tune. Like most methods that do not assume a statistical model of the data (including many density-based clustering methods), CLASSIX may suffer from a chaining effect where  poorly separated clusters inadvertently merge. In such cases, tuning the $\radius$ parameter for a balance between too many or too few clusters may be challenging. Similar issues have been observed with DBSCAN \cite{10.1145/2723372.2737792} and single linkage clustering methods. Perhaps a way to overcome this problem is to allow for the radius parameter to vary between different groups, with some adaptive method for choosing it. This may be a research problem for the future.}

Apart from the analysis of the simple Gaussian model in section~\ref{sec:simpmod}, we currently do not have a good theoretical understanding of why sorting helps even for high-dimensional ``real-world'' datasets such as those contained in the UCI Machine Learning Repository. While it is known that ``big data'' matrices often have low numerical rank \cite{udell2019big}, we usually find that the second singular value of the data matrix is still significant in comparison to the first. Hence, bounds like \eqref{eq:sigma2} cannot explain why data points can be aggregated with such a small number of pairwise distance computations as we have seen in the experiments. Further work will be needed to understand this inherent ``sortability'' of big data. 

\medskip

\noindent \textbf{Acknowledgement.} 
We are grateful to Kamil Oster who provided us with the dataset used in the Appendix. \rev{We also thank the two anonymous referees who have provided many insightful comments that have improved the manuscript.}

\appendix

\bibliographystyle{abbrv}

\bibliography{cas-refs.bib}

\end{document}